\pdfoutput=1
\documentclass[12pt,onecolumn]{article}
\usepackage[utf8]{inputenc}
\usepackage[left=1.5cm, right=1.5cm, top=1.5cm, bottom=1.5cm]{geometry}
\usepackage{amsmath,amsfonts,amssymb}
\usepackage{graphicx}
\usepackage{float}
\usepackage{afterpage}
\usepackage[super,nocompress]{cite}
\usepackage{xurl}
\usepackage{setspace}
\usepackage{times}
\usepackage{titlesec}
\usepackage{eso-pic}
\usepackage[colorlinks=true,linkcolor=blue,citecolor=blue,urlcolor=blue,hypertexnames=false]{hyperref}
\usepackage{caption}
\usepackage{tabularx}
\usepackage{booktabs}
\usepackage{xcolor}
\definecolor{figurerefblue}{RGB}{26,13,172}
\usepackage{longtable}
\usepackage{makecell}

\AtBeginDocument{%
  \hypersetup{%
    pdfborder={0 0 1},%
    pdfborderstyle={/S/U/W 0.8},%
    linkcolor=figurerefblue,%
    citecolor=figurerefblue,%
    linkbordercolor=figurerefblue,%
    citebordercolor=figurerefblue,%
    urlbordercolor=blue%
  }%
}

\newcolumntype{L}[1]{>{\raggedright\arraybackslash}p{#1}}
\newcolumntype{C}[1]{>{\centering\arraybackslash}p{#1}}
\newcommand{\figref}[2][]{\mbox{\hyperref[#2]{\textbf{Figure~\ref*{#2}#1}}}}
\newcommand{\suppfigref}[2][]{\mbox{\hyperref[#2]{\textbf{Supplementary Fig.~\ref*{#2}#1}}}}
\newcommand{\methodsref}[1]{\hyperref[#1]{Methods}}
\newcommand{\methodsanchor}[1]{\phantomsection\label{#1}}

\titleformat*{\section}{\Large\bfseries}
\titlespacing*{\section}{0pt}{3em}{0.2em}
\titleformat*{\subsection}{\large\bfseries}
\titlespacing*{\subsection}{0pt}{1.5em}{-1.0em}

\begin{document}
\fontsize{13}{15.6}\selectfont
\setstretch{1.0}
\thispagestyle{plain}

\noindent{\fontsize{20}{24}\selectfont\sffamily\bfseries Auditable CT phenotyping through report-derived radiological observations\par}
\vspace{1.1em}
{\setstretch{1.15}\fontsize{13.5}{16.5}\selectfont
\noindent
Riga Wu\textsuperscript{1},
Walter R. Witschey\textsuperscript{2},
Yicheng Li\textsuperscript{2},
Felix Barajas Ordonez\textsuperscript{3},
Keno K. Bressem\textsuperscript{4,5,6},
Lisa C. Adams\textsuperscript{7},
Gary E. Weissman\textsuperscript{8,9},
Li Shen\textsuperscript{9},
Christos Davatzikos\textsuperscript{1,10},
Eduardo Mortani Barbosa Jr\textsuperscript{2},
Daniel Truhn\textsuperscript{3,$\ddagger$},
\& Tianyu Han\textsuperscript{2,10,*,$\ddagger$}
\par}
\vspace{1.1em}

{\setlength{\parskip}{0pt}%
\fontsize{12}{14.2}\selectfont
\itshape
\raggedright
\noindent
\textsuperscript{1}Artificial Intelligence in Biomedical Imaging Laboratory (AIBIL), Perelman School of Medicine, University of Pennsylvania, Philadelphia, PA, USA.\\[0.05em]
\textsuperscript{2}Department of Radiology, Perelman School of Medicine, University of Pennsylvania, Philadelphia, PA, USA.\\[0.05em]
\textsuperscript{3}Department of Diagnostic and Interventional Radiology, University Hospital RWTH Aachen, Aachen, Germany.\\[0.05em]
\textsuperscript{4}Department of Diagnostic and Interventional Radiology, School of Medicine and Health, Technical University of Munich, Munich, Germany.\\[0.05em]
\textsuperscript{5}National Center for Tumor Diseases West, Essen, Germany.\\[0.05em]
\textsuperscript{6}Institute of Artificial Intelligence in Medicine, University Hospital Essen, Essen, Germany.\\[0.05em]
\textsuperscript{7}Department of Diagnostic and Interventional Radiology, School of Medicine and Health, Klinikum rechts der Isar, TUM University Hospital, Technical University of Munich, Munich, Germany.\\[0.05em]
\textsuperscript{8}Division of Pulmonary, Allergy and Critical Care, Department of Medicine, Perelman School of Medicine, University of Pennsylvania, Philadelphia, PA, USA.\\[0.05em]
\textsuperscript{9}Department of Biostatistics, Epidemiology and Informatics, Perelman School of Medicine, University of Pennsylvania, Philadelphia, PA, USA.\\[0.05em]
\textsuperscript{10}Center for Artificial Intelligence and Data Science for Integrated Diagnostics (AI2D), Perelman School of Medicine, University of Pennsylvania, Philadelphia, PA, USA.\par

\vspace{0.8em}
\noindent\textsuperscript{$\ddagger$}These authors jointly supervised this work: Daniel Truhn and Tianyu Han.\\[0.4em]
\noindent\textsuperscript{*}\textbf{Corresponding author:} Tianyu Han (tianyu.han@pennmedicine.upenn.edu)
\par
}

\vspace{1.1em}
{\fontsize{12}{14.5}\selectfont\bfseries\noindent
Medical image foundation models can predict clinical phenotypes from computed tomography (CT)\cite{hamamci2025generalist,blankemeier2024merlin,huang2023inspect}, but strong performance leaves open whether they read disease-specific findings or shortcuts that correlate with the diagnosis\cite{zech2018variable,badgeley2019hip,degrave2021shortcuts}.
We tested this in 221 electronic-health-record (EHR) phenotypes using Auditable CT phenotyping (ACT), built on report-derived radiological observations.
We trained ACT on 38,317 patients, mined 376,194 observations and evaluated it in 25,183 held-out patients.
ACT exceeded five vision-language baselines on zero-shot annotation, and CT-CLIP across 221 phenotypes from unseen CT pulmonary angiography, both under zero-shot scoring (0.651 versus 0.572) and under linear probing (0.709 versus 0.662).
Reading each probe exposes what accuracy conceals: only 97 observations occupy the 221 rank-1 positions, and one phrase describing aortic and coronary calcification ranks first for 20 phenotypes, including osteoporosis, urinary tract infection and major depressive disorder.
Restricting the bank to clinician-specified evidence redirects those probes onto phenotype-related observations in 86 phenotypes at no accuracy cost (0.751 versus 0.741).
Accurate CT-based EHR phenotyping can therefore rest on observations that are not valid evidence for the coded phenotype and that ACT can identify and intervene on. 
\par}

\newpage
\section*{Introduction}

Medical image-text foundation models now learn clinical representations from reports and scientific text.
They support zero-shot recognition, retrieval, question answering and generation across radiography, pathology and other biomedical images\cite{radford2021learning,zhang2022convirt,huang2021gloria,boecking2022biovil,tiu2022chexzero,wang2022medclip,zhang2025multimodal,li2023llavamed,moor2023medflamingo,tu2024generalist,zhang2024biomedgpt,lu2024visual,kim2024transparent}.
Related work has tested generalist medical-image interpretation, evaluated image-containing clinical vignettes and released open-source medical instruction-tuning models and data\cite{han2023generalist,han2024comparative,han2023medalpaca}.
Studies in 3D have extended this approach to complete computed tomography (CT) volumes\cite{hamamci2025generalist,blankemeier2024merlin,wu2025radfm}.
Merlin and INSPECT went further and linked such volumes to outcomes recorded in the electronic health record (EHR)\cite{blankemeier2024merlin,huang2023inspect}.
Accurate prediction, however, falls short of establishing that a model uses disease-specific radiological evidence.
Zech et al. found that pneumonia classifiers identified hospital origin in more than 99.9\% of radiographs\cite{zech2018variable}.
Badgeley et al. reported that a hip-fracture classifier's area under the receiver operating characteristic curve (AUROC) fell from 0.78 to 0.52 once patient and hospital-process variables were matched\cite{badgeley2019hip}.
DeGrave et al. showed that radiographic coronavirus disease 2019 (COVID-19) detectors selected confounders rather than pulmonary pathology\cite{degrave2021shortcuts}.
Seyyed-Kalantari et al. documented higher underdiagnosis for Hispanic women than white women in one chest-radiograph cohort\cite{seyyedkalantari2021underdiagnosis}.
Gichoya et al. predicted self-reported race with AUROCs from 0.91 to 0.99 in radiographs and from 0.87 to 0.96 in chest CT\cite{gichoya2022race}.
Other studies showed that sex imbalance shifted performance and that models encoded protected characteristics\cite{larrazabal2020gender,glocker2023algorithmic}.
A systematic review found that none of 62 fully evaluated COVID-19 imaging models was clinically usable, because of methodological flaws or bias\cite{roberts2021common}.
These failures exemplify shortcut learning and hidden stratification\cite{geirhos2020shortcut,oakdenrayner2020hidden}.
The risk grows for EHR phenotypes.
Raw records reflect both the patient's state and the process that recorded it, and code-based definitions buy scale at the cost of clinical specificity\cite{hripcsak2013next,denny2010phewas,denny2013systematic,kirby2016phekb,wei2017evaluating,verma2022penn}.
A high phenotype AUROC can therefore leave open whether a predictor reads the target pathology or a proxy for site, acquisition or care process.

Auditable CT phenotyping (ACT) makes report-derived radiological observations the common language for image annotation, phenotyping, model audit and restriction (\figref[d--f]{fig:overview}).
ACT combines a native volume-report model with an observation bank extracted from the same reports.
The model was trained on 38,317 patients from CT-RATE and Merlin, spanning non-contrast chest and contrast-enhanced abdominal CT\cite{hamamci2025generalist,blankemeier2024merlin}.
Every one of the $M=376{,}194$ atomic observations carries two representations.
The native 768-dimensional text encoder matches it to images, and F2LLM, a large language model (LLM) text encoder, gives it a 5,120-dimensional semantic direction\cite{zhang2026f2llm}.
For a given volume, ACT's native image-text similarities weight the corresponding F2LLM directions.
The normalized weighted sum of those directions is the concept-anchored CT embedding.
The complete mathematical definition is given in \methodsref{meth:concept-representation}.
We can therefore search ACT's concept-anchored CT representation with report-derived observation directions.
Where a conventional concept bottleneck model (CBM)\cite{koh2020concept} reserves one latent coordinate per named observation, ACT anchors observations as directions and recovers them by projection.

Earlier work formalized concept-level analysis through testing with concept activation vectors (TCAV), automatic concept discovery, concept whitening and completeness measures\cite{kim2018tcav,ghorbani2019ace,chen2020concept,yeh2020completeness}.
CBMs, concept embeddings and post-hoc or label-free variants made concepts predictive and editable, but they also exposed trade-offs in supervision and leakage\cite{koh2020concept,zarlenga2022concept,yuksekgonul2023posthoc,oikarinen2023labelfree,havasi2022leakage}.
CLIP-Dissect, physician-guided auditing and SemanticLens further scaled semantic labelling and model inspection\cite{oikarinen2023clipdissect,degrave2025auditing,dreyer2025semanticlens}.
ACT instead reuses one report-derived radiological vocabulary across annotation, phenotyping, audit and restriction.
ACT constructs concept-anchored projections over volumetric CT and tests them against linked EHR phenotypes through paired probe-observation audits.

\begin{figure}[!t]
    \centering
    \captionsetup{skip=4pt}
    \includegraphics[width=\textwidth]{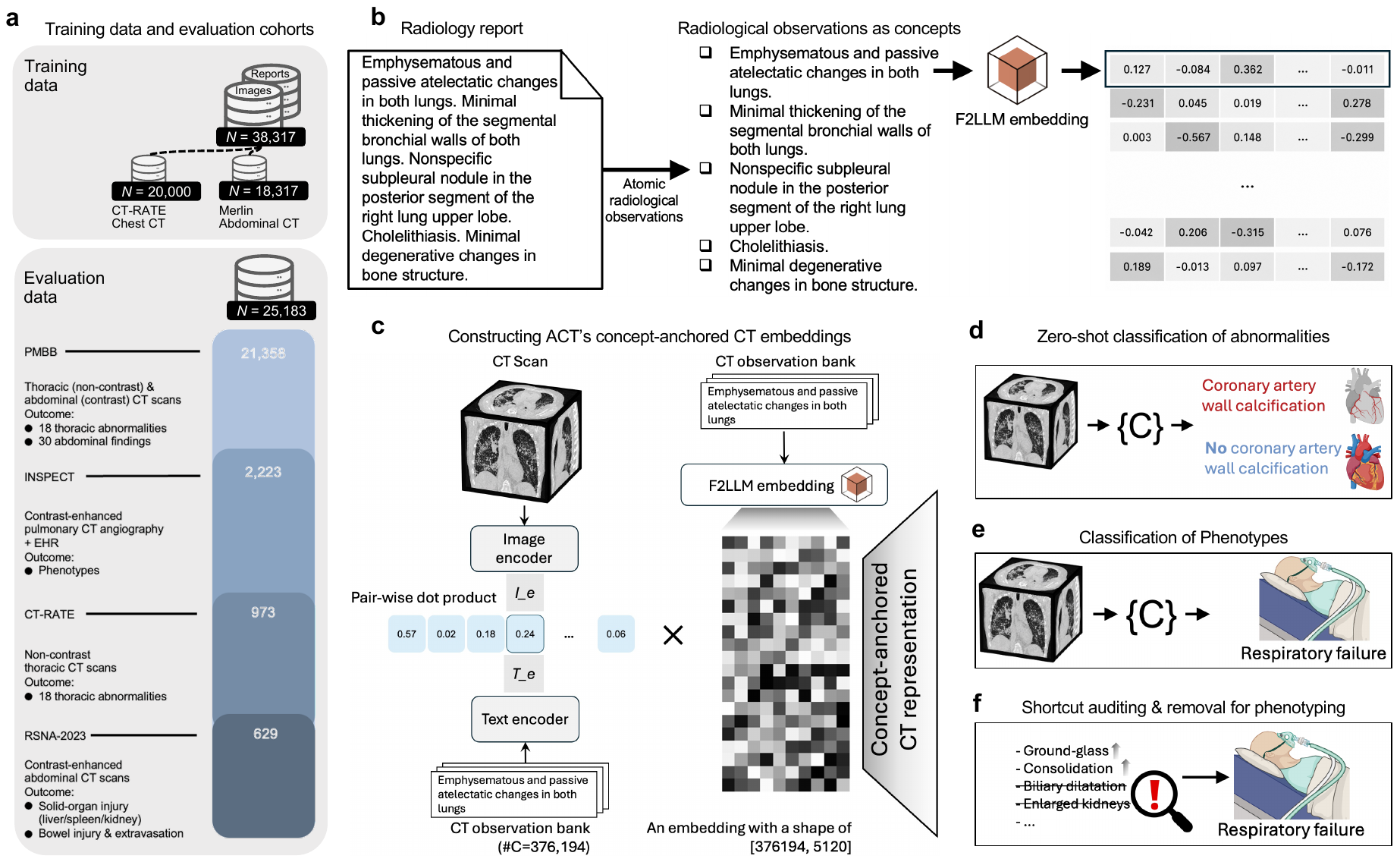}
    \caption{\textbf{ACT links report-derived observations to prediction and audit.}
    \textbf{a}, Training used chest CT from 20,000 CT-RATE patients and abdominal CT from 18,317 Merlin patients.
    PMBB, INSPECT, held-out CT-RATE and RSNA-2023 were reserved for evaluation.
    \textbf{b}, Reports were decomposed into 376,194 atomic radiological observations and embedded with F2LLM.
    \textbf{c}, For each CT, ACT's native image-text similarities weighted the corresponding F2LLM observation embeddings to yield ACT's normalized 5,120-dimensional concept-anchored CT representation.
    \textbf{d}--\textbf{f}, ACT's native model supported zero-shot abnormality annotation (\textbf{d}), whereas ACT's concept-anchored CT representation supported EHR phenotype classification (\textbf{e}) as well as probe-observation audit and observation-bank restriction (\textbf{f}).}
    \label{fig:overview}
\end{figure}

Existing approaches provide only parts of this chain.
MONET showed that concepts derived automatically from 105,550 dermatology image-text pairs can support dataset and model audits\cite{kim2024transparent}.
It stopped short of volumetric imaging and EHR phenotypes.
CT-CLIP, Merlin and RadFM learn from 3D scans and text, but leave untested whether EHR phenotype probes depend on phenotype-related, named observations\cite{hamamci2025generalist,blankemeier2024merlin,wu2025radfm}.
RadLex and the Unified Medical Language System (UMLS) organize radiological and biomedical terms and their relations\cite{langlotz2006radlex,bodenreider2004umls}.
SapBERT and MedCPT learn biomedical semantic representations\cite{liu2021sapbert,jin2023medcpt}.
None of them takes a single observation bank through the whole chain.
Such a chain would link the bank to CT volumes, validate it across anatomies and institutions, carry it into an unseen acquisition domain and then reuse it to inspect and restrict the predictor.
This study provides that missing end-to-end test.

We designed four linked tests to ask whether ACT's shared observation language is useful, organized and actionable.
These were zero-shot abnormality annotation (\figref[d]{fig:overview}), anatomical and ontological organization (\figref[b]{fig:overview}), transfer to 221 phenotypes derived from CT pulmonary angiography (CTPA) and the EHR (\figref[e]{fig:overview}), and audit plus clinical restriction (\figref[f]{fig:overview}).
We evaluated ACT in 25,183 held-out patients across CT-RATE, the Penn Medicine BioBank (PMBB), the Radiological Society of North America 2023 abdominal trauma dataset (RSNA-2023) and INSPECT (\figref[a]{fig:overview})\cite{hamamci2025generalist,verma2022penn,rsna2023abdominal,huang2023inspect}.
ACT's native 3D volume-report model achieved the highest mean abnormality-annotation AUROC among five complete-cohort 2D and 3D vision-language baselines in every evaluated cohort (\figref{fig:annotation}).
ACT's fixed observation bank was anatomically organized.
Among RadLex-grounded observations, Euclidean distance increased with ontology separation.
Observations from the external PMBB corpus fell within the corresponding regions of the reference map (\figref{fig:concept_space}).
In CTPA, which was absent from image pretraining, ACT's concept-anchored CT representation exceeded CT-CLIP in macro AUROC across 221 phenotypes, both under zero-shot scoring and under matched linear probing (\figref{fig:phenotyping}).
ACT's probe-observation audit exposed leading directions that lack evidence for the coded phenotype, including single directions shared across phenotypes with unrelated clinical targets (\figref{fig:proxy_profiles}).
ACT's phenotype-specific observation-bank restriction redirected selected probes toward phenotype-related observations and achieved a mean held-out test AUROC of 0.751 across 86 eligible phenotypes (\figref{fig:clinical_prior_refinement}).
ACT therefore exposes the radiological observation directions associated with phenotype probes, and it can restrict them.

\section*{Results}

\subsection*{ACT's native 3D model annotates concepts across diverse CT cohorts}

ACT couples a native 3D volume-report model to a bank of named radiological observations.
We first evaluated ACT's native model as an automatic concept annotator\cite{kim2024transparent,lu2024visual} before applying ACT's concept-anchored CT representation to phenotype prediction.
ACT's native model encodes each axial slice with a DINOv2 image encoder\cite{oquab2023dinov2,darcet2024vision} and aggregates the slice sequence into one volume embedding through a lightweight Transformer.
A contrastive objective then aligns that embedding to the free text of the paired report\cite{radford2021learning} (\suppfigref{fig:arch} and \methodsref{meth:native-model}).
We trained it on paired CT volumes and radiology reports from 38,317 patients, namely 20,000 with non-contrast chest CT from CT-RATE\cite{hamamci2025generalist} and 18,317 with contrast-enhanced abdominal CT from Merlin\cite{blankemeier2024merlin}.
To construct ACT's observation bank, we mined 376,194 distinct radiological observations from the same reports.
Each was embedded with F2LLM\cite{zhang2026f2llm}, a recent LLM text-embedding model that maps text to a 5,120-dimensional semantic vector.
Any volume can then be described by its similarity to this bank of named findings (\figref[b,c]{fig:overview}).
We evaluated ACT's native model on three held-out cohorts that span chest and abdomen as well as contrast and non-contrast acquisition.
These were CT-RATE (2,489 scans from 973 patients, an in-domain split), the external PMBB cohort (23,387 scans from 21,358 patients)\cite{verma2022penn} and the RSNA-2023 abdominal trauma dataset (943 scans in 629 reconstructed patient clusters)\cite{rsna2023abdominal}.
The PMBB scans comprised 9,097 chest and 14,290 abdominal examinations, and 2,029 patients were represented in both regional pools.
The annotation and retrieval analyses treated chest and abdominal PMBB separately, giving four evaluation cohorts.
We scored CT-RATE and RSNA-2023 against each dataset's reference labels, and PMBB against labels mined from its reports.
A further held-out dataset, INSPECT\cite{huang2023inspect}, comprised 2,612 phenotype-labelled CTPA scans from 2,223 patients and was reserved for the phenotyping experiments that follow.

\begin{figure}[p]
    \centering
    \includegraphics[width=\textwidth,height=\textheight,keepaspectratio]{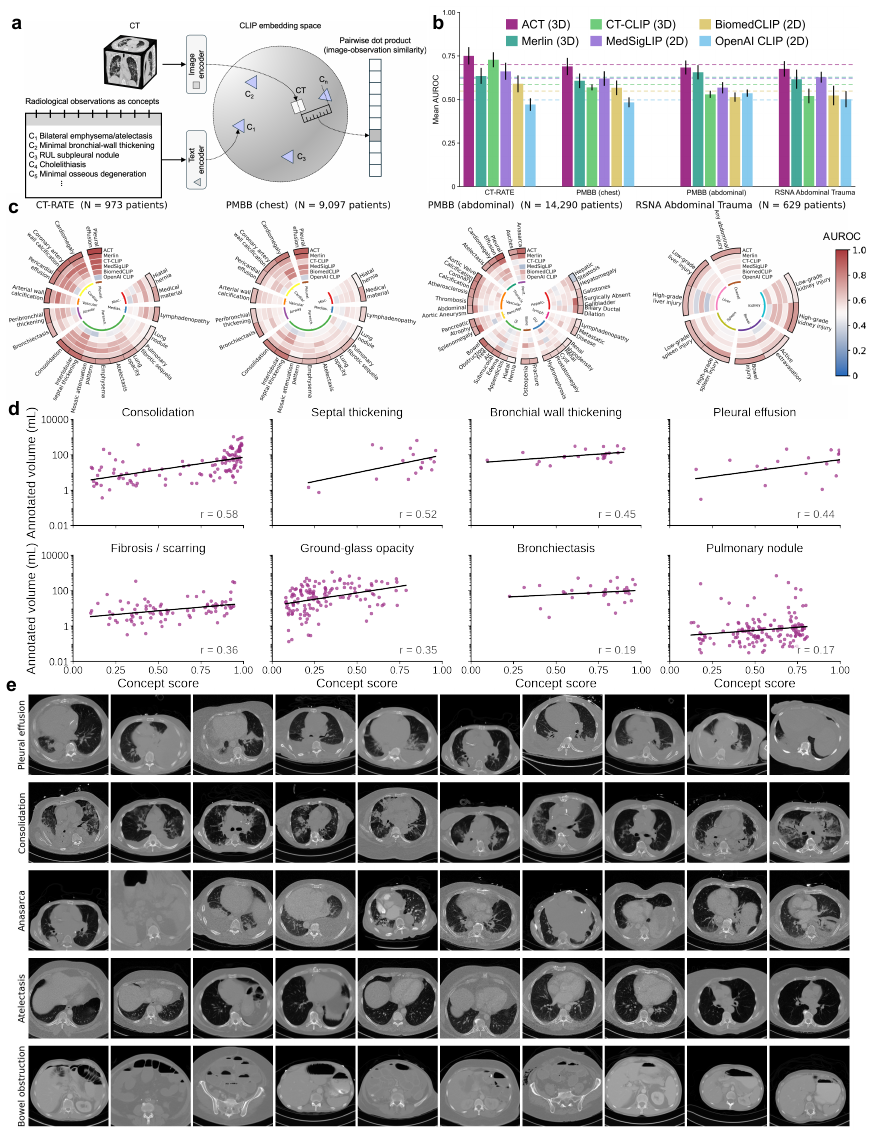}
\end{figure}

\afterpage{%
\captionof{figure}{\textbf{ACT's native 3D model annotates concepts across diverse CT cohorts.}
    \textbf{a}, Zero-shot finding scores were computed from image similarities to finding-presence and finding-absence prompts.
    \textbf{b}, Mean AUROC across findings for ACT's native model and five baselines.
    Data are the mean $\pm$1.96 times the standard error of the mean (s.e.m.) across finding-level AUROCs, and dashed lines show each model's mean across the four cohorts.
    \textbf{c}, Per-finding AUROC grouped by project-defined anatomical sectors.
    Rings denote models and colour denotes AUROC.
    Finding order was set on the reference cohort and reused.
    Cohorts contained 973 CT-RATE, 9,097 chest PMBB and 14,290 abdominal PMBB patients and 629 RSNA-2023 patient clusters.
    \textbf{d}, Concept scores scale with expert-annotated lesion volume in ReXGroundingCT\cite{baharoon2026rexgroundingct}.
    Each panel plots the concept-presence score for one finding against the volume of its expert segmentation, in the 313 held-out test volumes (283 patients) that carry such an annotation.
    Each point is one scan, volumes are shown on a logarithmic scale and $r$ denotes the Pearson correlation.
    \textbf{e}, Ten held-out PMBB volumes for each of five findings.
    Each volume is shown at the axial slice with the largest positive attribution computed from gradients and activations.
    \label{fig:annotation}}
\vspace{0em}
\noindent\rule{\textwidth}{0.4pt}
\vspace{0em}}

Across these cohorts, ACT's native model annotated the labelled findings zero-shot, comparing each volume's embedding with text prompts for the presence and the absence of a finding.
It attained the highest mean AUROC among five 2D and 3D vision-language baselines, each providing scores for every scan and finding in each cohort (\figref[a,b]{fig:annotation}).
It reached a mean AUROC of 0.749 on CT-RATE (18 thoracic findings), 0.689 and 0.683 on the chest and abdominal PMBB cohorts (18 thoracic and 30 abdominal findings, respectively), and 0.675 on RSNA-2023 abdominal trauma (nine abdominal-trauma outcomes).
These values exceeded both 3D baselines (CT-CLIP\cite{hamamci2025generalist} at 0.727 on CT-RATE and the abdomen-pretrained Merlin\cite{blankemeier2024merlin} at 0.656 on abdominal PMBB) and all three 2D baselines: MedSigLIP\cite{sellergren2025medgemma}, BiomedCLIP\cite{zhang2025multimodal} and OpenAI CLIP\cite{radford2021learning}.
ACT's native model achieved the highest per-finding AUROC on most findings in each cohort.
It won 13 of 18 findings on CT-RATE, 16 of 18 on chest PMBB, 6 of 9 on RSNA-2023 and 18 of 30 on abdominal PMBB (\figref[c]{fig:annotation}).
Supplementary Tables~\ref{tab:supp-zs-ctrate-test-a}--\ref{tab:supp-zs-rsna2023-test-b} report full per-finding AUROCs and patient-clustered 95\% confidence intervals (CIs) for CT-RATE, chest PMBB, abdominal PMBB and RSNA-2023.

The largest margins followed a consistent clinical pattern.
ACT's native model was strongest on abnormalities defined by distributed parenchymal or airway texture and on acute, emergent processes.
In both, the evidence is spread across many slices rather than confined to a single organ.
Consolidation was the clearest case, reaching an AUROC of 0.86 on CT-RATE and 0.81 on chest PMBB, gains of 0.13 and 0.12 over the strongest baseline in each cohort.
Interlobular septal thickening showed smaller gains of 0.08 and 0.09 in the same two cohorts.
The same pattern held in the acute abdomen.
Detection of bowel obstruction rose to an AUROC of 0.85, an improvement of 0.27 over the best baseline and the widest margin of any finding.
Free air reached 0.77, and on RSNA-2023 any abdominal injury reached 0.73, each at least 0.12 above the strongest competitor.
ACT's native model also led on appendicitis and traumatic bowel injury.
Baselines were otherwise most competitive on abdominal findings.
Merlin, which was pretrained on abdominal CT, remained the strongest there and kept an edge on organ-size and organ-specific density findings such as splenomegaly, hepatomegaly and hepatic steatosis.
This division is consistent with the models' different pretraining anatomies.
It suggests that ACT's native volume-report model was particularly effective for diffuse or acute findings.

The concept scores also tracked disease extent.
In ReXGroundingCT scans with voxel-level annotations\cite{baharoon2026rexgroundingct}, the zero-shot score correlated with the annotated volume of spatially extensive findings, reaching Pearson correlations of 0.58 for consolidation and 0.52 for septal thickening (\figref[d]{fig:annotation}).
Correlations were weakest for focal or airway-defined findings, at 0.17 for pulmonary nodules and 0.19 for bronchiectasis.

We next used concept-conditioned retrieval to test whether the concept scores could also rank images (\suppfigref{fig:retrieval}).
We ranked CT volumes by a queried finding within 64-candidate pools and computed 95\% CIs with a patient-clustered bootstrap\cite{zhang2025multimodal,huang2023visual}.
Random ranking gives floors of 1.6\%, 7.8\% and 15.6\% at Recall@1, @5 and @10.
ACT's native model achieved the highest Recall@10 on all four cohorts, reaching 46.9\% on CT-RATE (versus 43.0\% for CT-CLIP and 32.8\% for Merlin), 38.6\% on chest PMBB (versus 29.8\% for the next-best model, MedSigLIP), 37.3\% on abdominal PMBB and 34.1\% on RSNA-2023 (versus 27.5\% for Merlin).
It also outperformed every baseline at Recall@5 in every cohort except abdominal PMBB.
Each of those advantages had a paired CI that excluded zero.

The high-score montage showed qualitatively which cases the concept ranking retrieved.
For pleural effusion, consolidation, anasarca, atelectasis and bowel obstruction, we displayed ten high-scoring held-out PMBB volumes.
We represented each volume by the axial slice with the largest positive attribution computed from gradients and activations.
The displayed slices appeared compatible with the queried finding (\figref[e]{fig:annotation}).
Cross-institutional performance loss has been documented in medical imaging\cite{albadawy2018crossinstitutional}.
The zero-shot annotation and retrieval results on PMBB and RSNA-2023 support transfer to these two external cohorts.

\subsection*{ACT's radiological observations form an anatomically structured reference space}

The breadth of ACT's radiological observation bank is essential to the analyses that follow.
It spans disease-defining, incidental and cross-anatomical observations, including pleural and pericardial effusions, vascular calcification and biliary dilatation.
We visualized the lexical content of the 16 largest keyword-assigned categories, which contained 324,983 observations, 86.4\% of ACT's observation bank (\suppfigref{fig:concept_wordclouds}).
Prior work has judged learned reference spaces by their agreement with an external ontology and by where new entities fall in an unchanged atlas\cite{rosen2026universal}.
Structured semantic spaces have also let researchers search and audit learned model components in human-readable terms\cite{dreyer2025semanticlens}.
We therefore asked whether ACT's radiological observation space forms a clinically organized reference atlas rather than an unstructured list of report phrases.
We tested three complementary properties: anatomical organization, correspondence with an external radiological ontology and placement of observations from an external institution into the fixed reference space.
The 2D visualization used uniform manifold approximation and projection (UMAP).

\begin{figure}[p]
    \centering
    \includegraphics[width=\textwidth,height=\textheight,keepaspectratio]{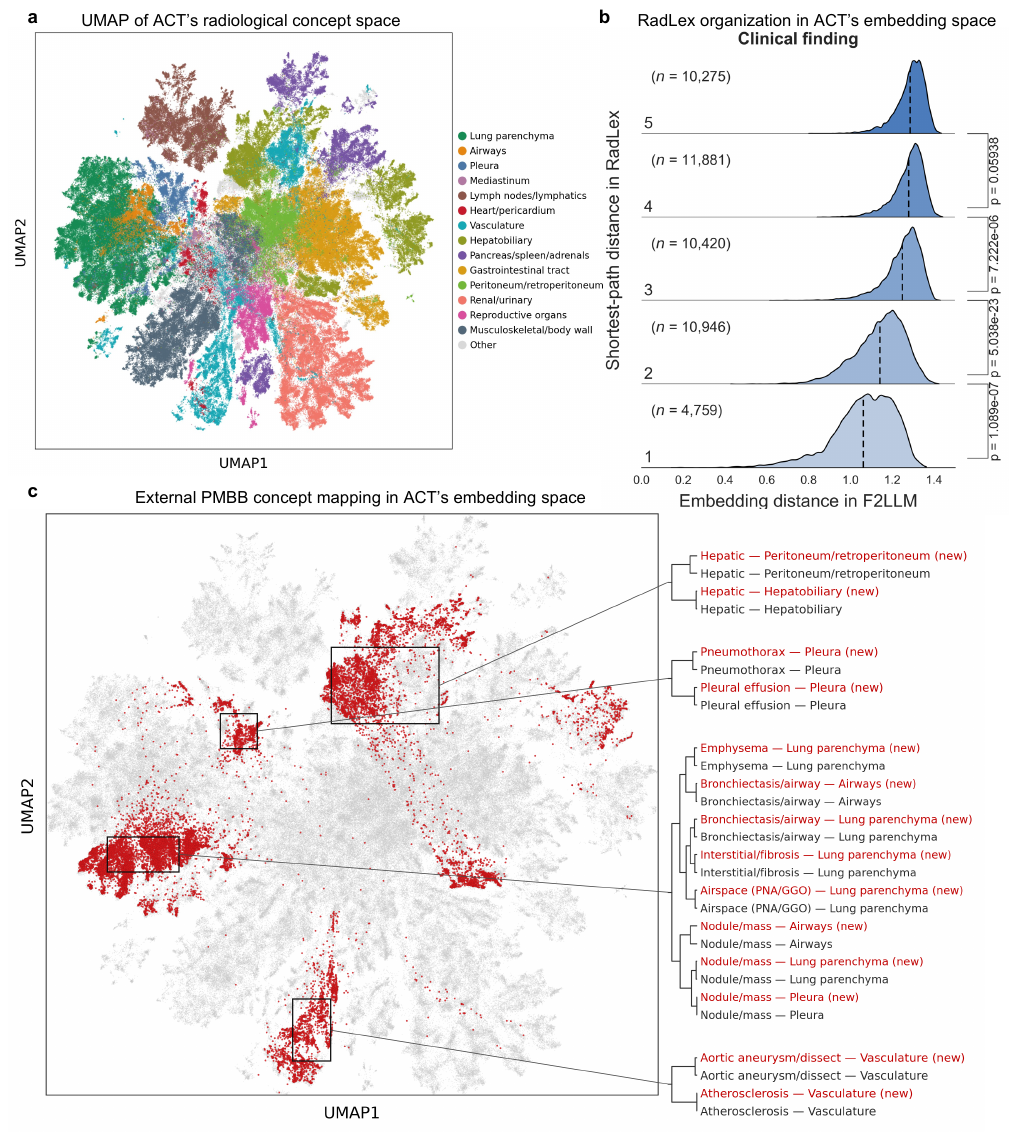}
\end{figure}

\afterpage{%
\captionof{figure}{\textbf{ACT's radiological observations form an anatomically structured reference space.}
    \textbf{a}, UMAP of all 376,194 F2LLM observation embeddings, coloured by 14 project-defined, RadLex-4.3-anchored anatomical families plus Other.
    \textbf{b}, Euclidean distances between $\ell_2$-normalized vectors in the original 5,120-dimensional space for 48,281 observation pairs spanning 883 pairs of 361 conservatively matched RadLex Clinical finding identifiers.
    Dashed lines mark means and $n$ denotes observation-pair counts.
    Distance tracked ontology separation (Spearman $\rho = 0.565$).
    Adjacent-hop one-sided Welch tests used identifier-pair means with Holm adjustment, and the four-to-five-hop increase remained unsupported ($p = 0.059$).
    \textbf{c}, Fixed-reference reconstruction of 20,751 strict PMBB-only observations (red) within the unchanged atlas (grey), from four prespecified finding-by-anatomy families: nodule or mass in lung parenchyma, hepatic findings in the hepatobiliary system, atherosclerosis in vasculature and pleural effusion in pleura.
    Coordinates were reconstructed from fixed reference neighbours without refitting.
    Boxes and rooted hierarchy excerpts summarize each family, and red leaf text marks PMBB-side entries rather than highlighted points.
    GGO, ground-glass opacity. PNA, pneumonia.
    \label{fig:concept_space}}
\vspace{0em}
\noindent\rule{\textwidth}{0.4pt}
\vspace{0em}}

The UMAP visualization\cite{mcinnes2018umap} of all 376,194 F2LLM observation embeddings showed broad, partially overlapping anatomical organization across 14 prespecified RadLex-4.3-anchored families (\figref[a]{fig:concept_space})\cite{langlotz2006radlex,rubin2008terminology,rsna2025radlex}.
Because UMAP is only a visualization, we next measured Euclidean distances between $\ell_2$-normalized vectors in the full 5,120-dimensional space.
To test that space against an external semantic hierarchy, we used conservative lexical grounding to match 20,307 observations to 361 identifiers in the RadLex Clinical finding branch.
Across 48,281 sampled phrase pairs spanning 883 RadLex-identifier pairs, Euclidean distance increased with RadLex shortest-path distance (Spearman $\rho=0.565$), from a mean of 1.061 at one hop to 1.285 at five hops (\figref[b]{fig:concept_space}).
Our criterion was ontology path length rather than proximity in the 2D plot.
Path length is an established basis for biomedical semantic-similarity measures\cite{pesquita2009semantic}.
Holm-adjusted one-sided Welch tests on identifier-pair means supported increases from one to two hops ($p=1.09\times10^{-7}$), two to three hops ($p=5.04\times10^{-23}$) and three to four hops ($p=7.22\times10^{-6}$), whereas the four-to-five-hop increment plateaued ($p=0.059$).

Finally, we embedded 157,466 unique observations from a separate PMBB extraction that covered all 9,097 chest reports and 7,799 of 14,290 abdominal reports, 16,896 reports in total.
We then placed them relative to the unchanged CT-RATE and Merlin atlas by fixed-reference neighbour reconstruction, without refitting the atlas (\figref[c]{fig:concept_space}).
The figure highlights 20,751 observations in four prespecified finding-by-anatomy families, namely nodule or mass in lung parenchyma, hepatic findings in the hepatobiliary system, atherosclerosis in vasculature and pleural effusion in pleura.
Qualitatively, these fell in the corresponding regions of the reference space.
In each of the four families, the PMBB and reference centroids were direct neighbours in the shared hierarchy.
Supplementary Table~\ref{tab:pmbb-polarity-neighbors} reports full-space examples for eight PMBB observations, with three same-finding affirmative and three explicitly negated neighbours from each reference corpus.
Together with the annotation experiments, these results support ACT's radiological observation space as a fixed semantic coordinate system for downstream phenotyping and auditing.

\subsection*{ACT's concept-anchored CT representation supports CTPA-EHR phenotyping beyond image pretraining}

We next tested ACT's concept-anchored CT representation in CTPA, an acquisition domain absent from image pretraining.
From INSPECT's linked EHR\cite{huang2023inspect}, we mapped diagnosis codes to 1,692 phecodes using the phenome-wide association study (PheWAS) framework\cite{denny2010phewas,wei2017evaluating,wu2019mapping}.
We retained the 221 phenotypes with at least 50 positive scans in the held-out test split.
Evaluation used its 2,612 CTPA scans from 2,223 patients.
For each scan, we compared the native CT-CLIP embedding\cite{hamamci2025generalist} with ACT's F2LLM-based concept-anchored CT embedding in two regimes.
The first was zero-shot scoring, which ran without INSPECT phenotype labels.
The second was supervised linear probing.
We fitted the linear probes to each frozen representation using the same INSPECT train, validation and test partitions, with learning rates selected on validation for each model.
Linear probing shows what information is accessible without changing the encoder.
Under domain shift, full fine-tuning can distort pretrained features and fall behind frozen-feature probes out of distribution\cite{kumar2022finetuning}.
The linear-probe experiment therefore measured how well representations that never saw CTPA adapt under supervision.

ACT's F2LLM-based concept-anchored CT representation had higher test classification performance than CT-CLIP in both regimes (\figref{fig:phenotyping}).
Macro AUROC was 0.651 (95\% CI, 0.638--0.662) versus 0.572 (95\% CI, 0.559--0.587) in zero-shot scoring and 0.709 (95\% CI, 0.699--0.718) versus 0.662 (95\% CI, 0.653--0.671) with linear probing.
On 1,000 shared patient-clustered bootstrap resamples, the paired macro-AUROC difference was 0.079 (95\% CI, 0.061--0.095) in zero-shot scoring and 0.047 (95\% CI, 0.036--0.056) with linear probing, and both intervals excluded zero.
Supplementary Table~\ref{tab:supp-linear-probe-221} reports the matched linear-probe AUROC for each of the 221 phenotypes, with two-sided 95\% CIs across fits.
Per-phenotype AUROC was descriptively higher for this representation in 182 of 221 phenotypes under zero-shot scoring and 190 of 221 with linear probing, including 171 in both regimes.
With linear probing, mean AUROC reached at least 0.75 for 71 phenotypes and at least 0.80 for 22, versus 17 and 3 for CT-CLIP.

\begin{figure}[p]
    \centering
    \includegraphics[width=\textwidth,height=\textheight,keepaspectratio]{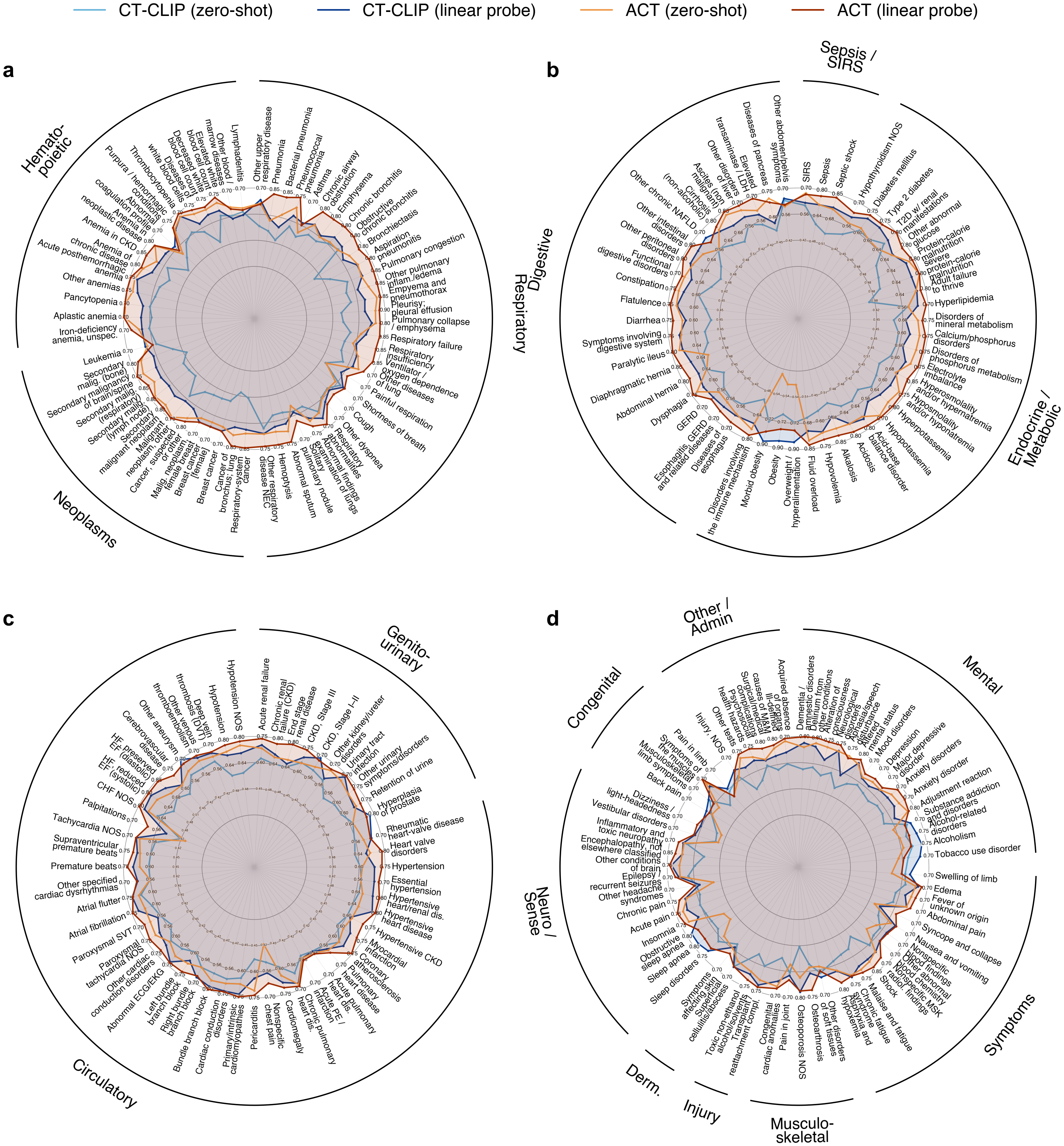}
\end{figure}

\afterpage{%
\captionof{figure}{\textbf{ACT's concept-anchored CT representation supports CTPA-EHR phenotyping beyond image pretraining.}
    Held-out test AUROC for 221 INSPECT EHR-derived phenotypes with at least 50 positive scans.
    CT-CLIP and ACT's concept-anchored CT representation were compared in 2,612 CTPA scans from 2,223 patients under zero-shot scoring and matched linear probing.
    CTPA was absent from image pretraining.
    \textbf{a}, Respiratory, neoplastic and hematopoietic phenotypes.
    \textbf{b}, Sepsis/SIRS, endocrine/metabolic and digestive phenotypes.
    \textbf{c}, Genitourinary and circulatory phenotypes.
    \textbf{d}, Mental, symptom-based and neurological phenotypes together with the remaining smaller sectors.
    Linear-probe curves are means across 20 matched probe fits.
    Patient-clustered bootstrap CIs for the overall comparison are given in the text.
    The nonclinical Other/Admin sector is placed last, and phecode order is retained within sectors.
    CKD, chronic kidney disease. GERD, gastro-oesophageal reflux disease. LDH, lactate dehydrogenase. MSK, musculoskeletal. NAFLD, non-alcoholic fatty liver disease. NEC, not elsewhere classified. NOS, not otherwise specified. SIRS, systemic inflammatory response syndrome. T2D, type 2 diabetes.
    \label{fig:phenotyping}}
\vspace{0em}
\noindent\rule{\textwidth}{0.4pt}
\vspace{0em}}

These differences between ACT and CT-CLIP were broad but heterogeneous across clinical sectors.
Mean advantages of ACT's concept-anchored CT representation over CT-CLIP were largest for respiratory phenotypes ($\Delta\mathrm{AUROC}=0.081$, higher for 28 of 30 phenotypes), neoplastic phenotypes (0.073, higher for 12 of 13) and hematopoietic phenotypes (0.056, higher for 15 of 16) (\figref[a]{fig:phenotyping}).
In both evaluation regimes, the concept-anchored CT representation had the higher AUROC for 25 of 30, 12 of 13 and 15 of 16 phenotypes in these sectors, respectively.
Positive mean advantages extended to sepsis/SIRS ($\Delta\mathrm{AUROC}=0.056$, higher for 3 of 3 phenotypes), endocrine/metabolic (0.052, higher for 23 of 26), digestive (0.051, higher for 19 of 21), genitourinary (0.039, higher for 8 of 10) and circulatory phenotypes (0.036, higher for 37 of 41) (\figref[b,c]{fig:phenotyping}).
Across the remaining clinical sectors in panel \textbf{d}, mean differences were also positive, except in the small congenital group.
There, mean linear-probe AUROC was 0.013 lower and only 1 of 4 phenotypes was higher (\figref[d]{fig:phenotyping}).
Because the representations differ in model provenance and dimensionality, these comparisons leave concept anchoring as one explanation among several for the performance difference.
They do establish that ACT's concept-anchored CT representation carries phenotype-relevant information in CTPA under both zero-shot and supervised evaluation.
They leave open which evidence supported each prediction.
Because EHR phenotypes can correlate with comorbid and incidental imaging findings, we next examined which named observations aligned with the learned phenotype directions.

\subsection*{ACT's probe-observation audit reveals candidate proxy directions}

A high held-out phenotype AUROC falls short of establishing that a probe relies on disease-specific imaging evidence.
Radiological models can keep their apparent classification performance while exploiting hospital source, care processes or image confounders\cite{zech2018variable,badgeley2019hip,degrave2021shortcuts,han2024confounders}.
Concept activation and semantic auditing methods expose such behaviour as human-readable directions or clinical hypotheses\cite{kim2018tcav,kim2024transparent,degrave2025auditing,dreyer2025semanticlens}.
We therefore applied ACT's probe-observation audit to ask which named radiological observation directions aligned with each fitted phenotype probe.
Using bank-mean-centred, $\ell_2$-normalized F2LLM observation vectors, we computed each phenotype probe's observation-alignment scores, averaged them across 20 probe fits and ranked all 376,194 observations.
The complete score definition is given in \methodsref{meth:probe-audit}.
For each of eight illustrative phenotypes (mean held-out test AUROC, 0.769 to 0.819), the five highest-ranking observations described one dominant observation family
(\figref{fig:proxy_profiles}).

\begin{figure}[p]
    \centering
    \includegraphics[width=\textwidth,height=\textheight,keepaspectratio]{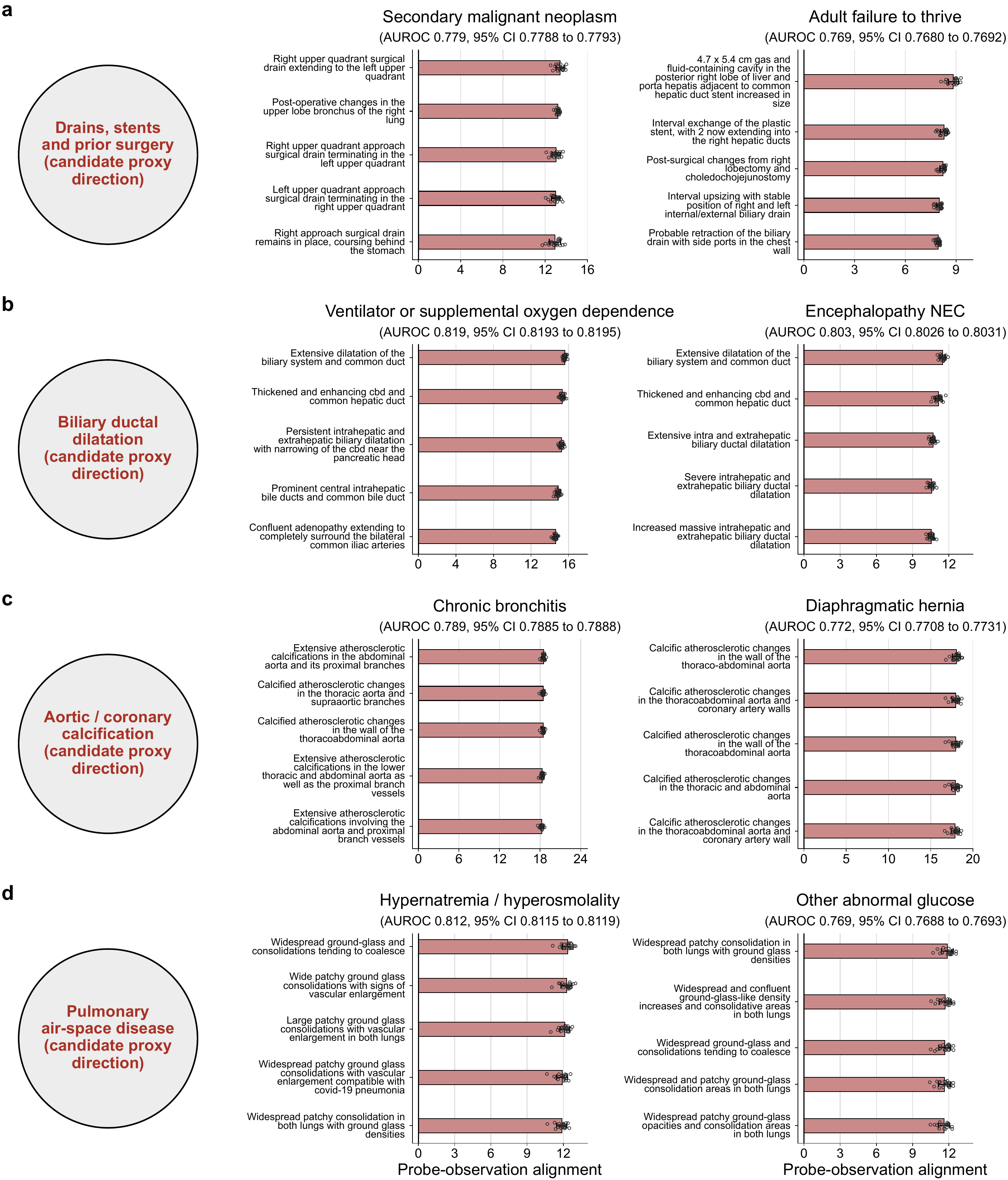}
\end{figure}

\afterpage{%
\captionof{figure}{\textbf{Candidate proxy directions from four observation families.}
    Each panel is one observation family, named in the circle at the left, and holds the five leading directions of two phenotypes drawn from unrelated three-digit phecode groups.
    \textbf{a}, Drains, stents and prior surgery, for secondary malignant neoplasm (left) and adult failure to thrive (right).
    \textbf{b}, Biliary ductal dilatation, for ventilator or supplemental oxygen dependence and encephalopathy NEC.
    \textbf{c}, Aortic or coronary calcification, for chronic bronchitis and diaphragmatic hernia.
    \textbf{d}, Pulmonary air-space disease, for hypernatremia or hyperosmolality and other abnormal glucose.
    The same string ranks first in both profiles of \textbf{b} and ranks first for 15 of the 221 phenotypes.
    Bars show mean projections across 20 fits, data are the mean $\pm$1 s.d., open circles show individual fits and title intervals are two-sided 95\% Student $t$ CIs across fits.
    \suppfigref{fig:shortcut_audit} shows the 13 further phenotypes that share the leading direction of both profiles in \textbf{b}. NEC, not elsewhere classified.
    \label{fig:proxy_profiles}}
\vspace{0em}
\noindent\rule{\textwidth}{0.4pt}
\vspace{0em}}

A single observation string was the highest-ranked direction for many phenotypes at once.
The union of all 221 top-5 lists contained only 389 distinct observations from the 376,194-observation bank, and just 97 strings filled the 221 rank-1 positions.
The phrase ``calcific atherosclerotic changes in the thoracoabdominal aorta and coronary artery walls'' ranked first for 20 phenotypes, among them osteoporosis not otherwise specified (NOS), urinary tract infection, chronic pain, left bundle branch block and major depressive disorder (\suppfigref[a]{fig:shortcut_audit}).
Those 20 phenotypes drew on only 22 distinct strings across their 100 top-5 positions, and their top-25 lists overlapped by a median of 14 of 25.
Osteoporosis NOS and major depressive disorder shared 14 of their top 25.
The phrase ``extensive dilatation of the biliary system and common duct'' ranked first for 15 phenotypes, including ventilator or supplemental oxygen dependence, encephalopathy not elsewhere classified (NEC) and retention of urine (\suppfigref[b]{fig:shortcut_audit}).
A direction that ranks first for a skeletal diagnosis and for a urinary infection is not phenotype-specific evidence for either.
Individual pairings admit clinical readings, but each such reading fails to cover the other phenotypes that share the same direction at the same rank.
To complement these shared-direction examples, we selected four phenotypes from distinct three-digit phecode groups.
In each, all ten of the natural top-10 observations were classed as either direct target evidence or clinically related but non-defining or qualified context.
Bronchiectasis aligned directly with bronchiectatic morphology, and pneumonia aligned with nonspecific air-space disease.
The coronary atherosclerosis profile combined direct coronary calcification with related aortic atherosclerosis, and the pulmonary-collapse profile reflected the collapse branch of a composite label (\suppfigref[a--d]{fig:clinical_alignment_audit}).

Across the complete natural, non-de-duplicated top-25 profiles, vascular calcification or atherosclerosis formed a lexical majority for 97 of 221 probes and pulmonary air-space disease for 54.
Among the 71 probes with mean held-out test AUROC of at least 0.75, the corresponding counts were 26 and 20.
To show that target-mismatched directions appear even when classification performance is high, we selected eight profiles post hoc among these 71 probes, four from the two screened families and four from reading every leading direction above the threshold (\figref{fig:proxy_profiles}).
Vascular directions dominated chronic bronchitis and diaphragmatic hernia (\figref[c]{fig:proxy_profiles}) as well as osteoporosis (\suppfigref[b]{fig:refinement_higher_auroc_profiles}), and pulmonary air-space directions dominated hypernatremia or hyperosmolality and other abnormal glucose (\figref[d]{fig:proxy_profiles}), neither of which has any imaging manifestation.
Two further families fell outside the lexical screen.
Surgical drains, biliary stents and post-surgical change led the profiles for secondary malignant neoplasm and adult failure to thrive (\figref[a]{fig:proxy_profiles}), and all of these record treatment rather than disease.
Biliary ductal dilatation led the profiles for ventilator or supplemental oxygen dependence and encephalopathy NEC (\figref[b]{fig:proxy_profiles}), neither of which involves the biliary tree.
The four families are broadcast to different degrees.
The vascular and biliary strings in \suppfigref{fig:shortcut_audit} rank first for 20 and 15 phenotypes, and the most widely shared air-space string for 10.
The drain string leading secondary malignant neoplasm ranks first for only one other phenotype, secondary malignancy of lymph nodes, and the stent string leading adult failure to thrive for none.
\figref{fig:proxy_profiles} therefore spans both a direction shared across many unrelated codes and a direction confined to a few related ones.
These recurring directions suggest three candidate mechanisms: age-linked vascular or cardiometabolic burden, acute pulmonary illness, and prior procedures or care context, the last covering both the device and the biliary families.
They stop short of identifying the patient-level signal used for classification.

Probe-observation projection measures global semantic correspondence.
We interpret the target-mismatched themes as candidate proxy directions that require patient-level and intervention-based validation.
We next asked whether a clinical prior on ACT's radiological observation space could suppress candidate non-target directions while preserving held-out phenotype AUROC.

\subsection*{ACT's observation-bank restriction redirects selected probes toward phenotype-related observations}

The audit becomes actionable only if ACT's observation space can also change which observations a phenotype predictor may use.
CBMs and post-hoc or label-free variants make named concepts part of the predictive representation\cite{koh2020concept,yuksekgonul2023posthoc,oikarinen2023labelfree}.
Concept leakage shows why interventions in that representation need separate validation\cite{havasi2022leakage}.
We therefore asked whether restricting the bank to phenotype-specific observations could redirect a downstream probe.
We report held-out classification performance after the restriction.

Within ACT, we restricted each phenotype to a rule-defined set of direct or associated observations.
We then recomputed its weighted semantic representation from only those observations and refitted its probe in the resulting subspace.
This procedure left ACT's native image representation and radiological observation bank frozen while excluding observation directions outside each phenotype-specific rule set.
Its complete mathematical definition is given in \methodsref{meth:bank-restriction}.
\figref[a--h]{fig:clinical_prior_refinement} shows eight illustrative pulmonary, cardiac and haemodynamic examples.

Across all 86 eligible phenotypes, the mean reported held-out test AUROC was 0.751 (95\% CI, 0.742--0.760) for ACT's rule-restricted probes and 0.741 (95\% CI, 0.731--0.750) for the full-bank ACT reference.
Restricted-probe AUROC was higher in 55 phenotypes and lower in 31.
Supplementary Table~\ref{tab:supp-restriction-all86} reports complete phenotype-level full-bank and rule-restricted AUROCs with matched patient-clustered bootstrap CIs.
The restriction rules excluded mentions of procedures or hardware, including drains, catheters and stents.
Terms that name anatomy or pathology rather than a device were retained, and the complete list is given in \methodsref{meth:clinical-screen}.
Eight additional post-hoc paired audits compared the leading full-bank and refined observation profiles.
Four examples had higher refined AUROC (\suppfigref{fig:refinement_higher_auroc_profiles}) and four had lower (\suppfigref{fig:refinement_lower_auroc_profiles}).
The osteoporosis panel revisits a phenotype whose full-bank profile was led by the shared vascular direction of \suppfigref[a]{fig:shortcut_audit}.
It draws that profile beside the restricted one (\suppfigref[b]{fig:refinement_higher_auroc_profiles}).
In these selected profiles, the leading observations became more clinically coherent whether held-out AUROC rose or fell.

Within the eight displayed examples, the fitted restricted probes placed their largest weights on the observations that clinical practice associates with each phenotype.
For congestive heart failure (CHF) the restriction retained pleural effusion, the characteristic thoracic manifestation of the syndrome\cite{porcel2010pleural}.
For hypovolemia it retained a flat or collapsed inferior vena cava (IVC), a documented CT sign of shock in trauma\cite{kim2022flativc}, and for shock it retained active contrast extravasation (\figref[c,d]{fig:clinical_prior_refinement}).
For pericarditis it retained pericardial effusion, which the 2025 European Society of Cardiology (ESC) guidelines identify as the imaging manifestation of the syndrome\cite{schulzmenger2025esc} (\figref[h]{fig:clinical_prior_refinement}).
Panels \textbf{e} to \textbf{g} followed the same pattern, retaining bilateral pleural effusions, subpleural nodular consolidation, and pneumothorax or atelectasis (\figref[e--g]{fig:clinical_prior_refinement}).
They are the closest evidence the modality can supply, and they remain global directions rather than patient-level evidence.
The restriction preserved accuracy.
Pneumococcal pneumonia rose from 0.744 to 0.794, and the restricted probe was higher in seven of the eight displayed examples, with CHF similar at 0.790 versus 0.794 (\figref[a,b]{fig:clinical_prior_refinement}).
A median restricted bank held 1,744 observations, 0.46\% of the full bank, and hypovolemia reached 0.730 from 0.708 on 56 observations (Supplementary Table~\ref{tab:supp-restriction-all86}).
Restricting the observation bank therefore changes the semantic basis of a predictor without costing the classification performance that drove it.

\begin{figure}[p]
    \centering
    \captionsetup{skip=5pt}
    \includegraphics[width=\textwidth]{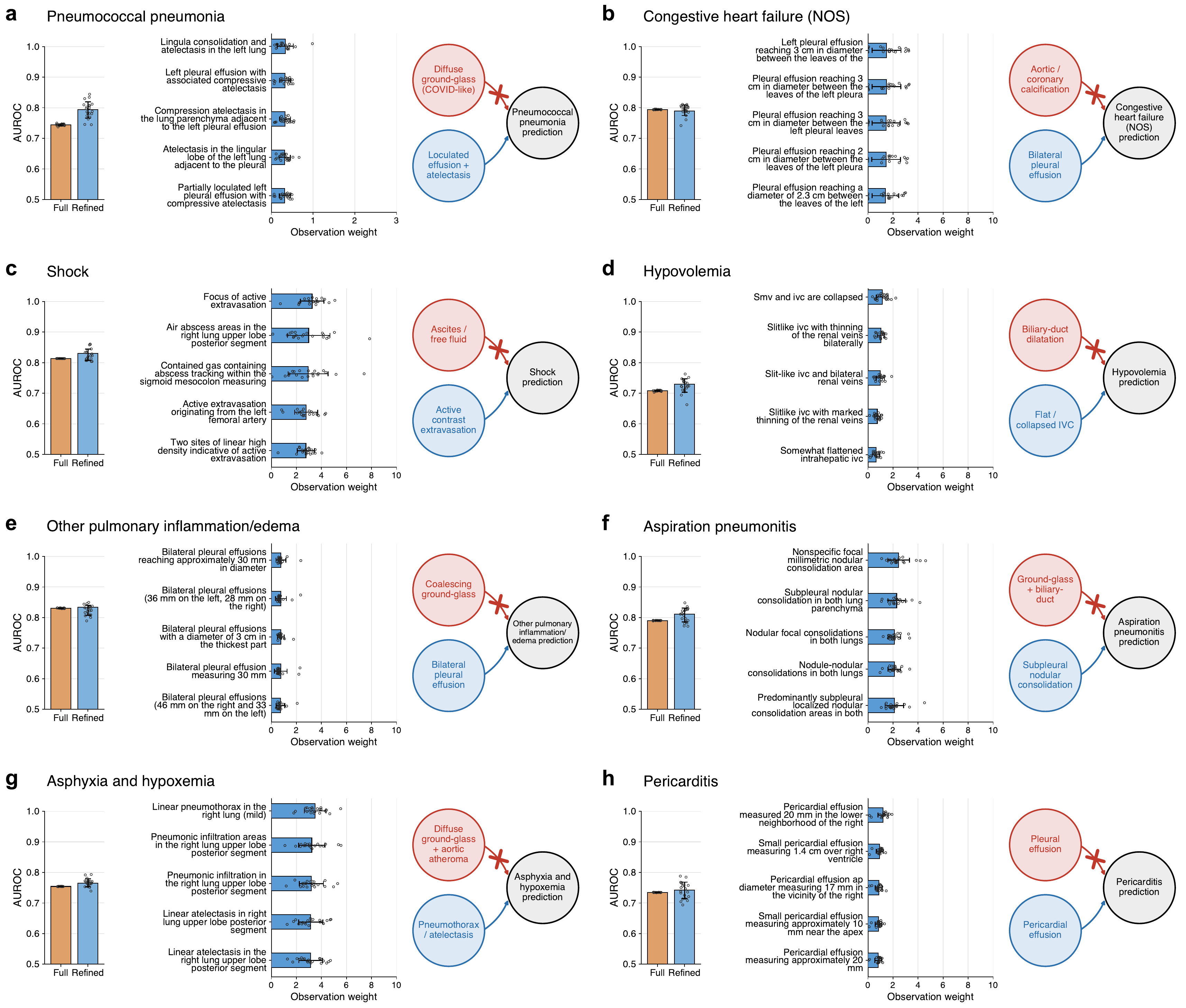}
    \caption{\textbf{ACT's observation-bank restriction redirects selected probes toward phenotype-related observations.}
    \textbf{a}--\textbf{h}, Each module shows held-out test AUROC (left), five retained positive-weight observations (middle) and a manual contrast between an excluded candidate non-target theme (red) and a retained phenotype-related theme (blue, right).
    The phenotypes are pneumococcal pneumonia (\textbf{a}), CHF (NOS) (\textbf{b}), shock (\textbf{c}), hypovolemia (\textbf{d}), other pulmonary inflammation or edema (\textbf{e}), aspiration pneumonitis (\textbf{f}), asphyxia and hypoxemia (\textbf{g}) and pericarditis (\textbf{h}).
    Full is the full-bank probe test AUROC, and Refined is the restricted probe test AUROC.
    Markers show individual estimates, and data are the mean $\pm$1 s.d.
    Retained observations were chosen from the fitted restricted probe's ten largest weights and ordered by mean projected weight across 20 training-set bootstrap refits.
    Displayed phrases retain near-duplicates, so a panel can show several wordings of one observation.
    In \textbf{b} the retained effusion phrases are all left-sided variants of one template, which reflects the phrasing cluster rather than a claim about side, and the restricted bank admits right-sided and bilateral effusion phrases in comparable numbers.
    Supplementary Table~\ref{tab:supp-restriction-all86} reports all 86 phenotype-level AUROCs and matched patient-clustered bootstrap CIs.
    CHF, congestive heart failure. IVC, inferior vena cava. NOS, not otherwise specified. SMV, superior mesenteric vein.}
    \label{fig:clinical_prior_refinement}
\end{figure}

\section*{Discussion}

In this study, we built ACT around one report-derived observation vocabulary.
We carried that vocabulary through image annotation, phenotype prediction, model audit and observation-bank restriction (\figref[d--f]{fig:overview}).
We trained ACT's native volume-report model on 38,317 patients with both non-contrast chest and contrast-enhanced abdominal CT, mined 376,194 observations from the same reports and evaluated ACT in 29,431 scans from 25,183 held-out patients in CT-RATE, PMBB, RSNA-2023 and INSPECT.
ACT's native model annotated 18 thoracic, 30 abdominal and nine trauma findings at the highest mean AUROC in every cohort, ahead of five 2D and 3D vision-language baselines.
In ACT's fixed observation space, embedding distance tracked separation in the RadLex ontology.
Observations from an external institution fell in the corresponding regions of an unchanged atlas.
In CTPA, which image pretraining never saw, ACT's concept-anchored CT representation exceeded CT-CLIP across 221 EHR-derived phenotypes under both zero-shot scoring and matched linear probing.
ACT's probe-observation audit then exposed leading directions that lack evidence for the coded phenotype, including one that ranked first for 20 of those phenotypes.
Restricting the bank under clinician-defined rules redirected selected probes toward phenotype-related observations in 86 eligible phenotypes, without costing held-out classification.
A phenotype predictor built this way can therefore be read and constrained in the words a radiologist writes.

CBMs reserve one latent coordinate per concept, so the vocabulary must be fixed and labelled before training\cite{koh2020concept}.
Post-hoc and label-free variants relax that requirement, yet leakage studies show a coordinate can encode something other than the concept it names\cite{yuksekgonul2023posthoc,oikarinen2023labelfree,havasi2022leakage}.
ACT anchors each observation as a direction in a fixed text-embedding space and recovers it by projection.
The bank can therefore span the full report vocabulary without concept annotation.
Tools that translate learned components into human-readable terms, among them TCAV, MONET and SemanticLens, stop at inspection\cite{kim2018tcav,kim2024transparent,dreyer2025semanticlens}.
None operates on CT volumes with linked EHR phenotypes, and none constrains what the predictor may use.

The largest annotation margins fell in a consistent place, namely findings whose evidence is distributed across the volume rather than concentrated in one structure.
Consolidation, interlobular septal thickening, bowel obstruction and free air are of this kind.
A model that fuses the whole slice sequence can integrate evidence spread beyond any single slice.
The abdomen-pretrained baseline instead kept its edge on organ-size and organ-specific density findings such as splenomegaly and hepatic steatosis.
These are released systems with different encoders, embedding dimensions, pretraining corpora and prompt templates, so the comparison leaves open whether the advantage came from the 3D encoder, the data or report supervision.

We judged the observation space by ontology path length in the full 5,120-dimensional embedding rather than by proximity in the UMAP plot.
The PMBB observations then landed in the expected regions, and the reference coordinates stayed as they were.
A reference space earns the name coordinate system only when it survives a new corpus intact\cite{rosen2026universal}.
The audit rests on that frame, because a probe direction is readable only if the names sit where a radiologist would put them.

The CTPA advantage reached well beyond respiratory phenotypes.
It extended across the neoplastic, hematopoietic, endocrine, digestive and circulatory sectors, and the small congenital group was the only exception.
The report vocabulary carried the transfer, because the native encoder met CTPA for the first time at evaluation and stayed frozen throughout.
The two representations differ in provenance and dimensionality, so concept anchoring remains one explanation among several.
Phecodes aggregate encounter diagnosis codes rather than adjudicated findings, so a scan may carry the target pathology, its complications, its treatment or the care process that generated it\cite{denny2010phewas,wei2017evaluating}.
A higher phenotype AUROC therefore sharpens a question that lies beyond it, namely which radiological evidence the probe reads.

Only 97 distinct strings occupied the leading positions of all 221 phenotype probes.
A phrase describing aortic and coronary calcification led 20 of those probes, including osteoporosis, urinary tract infection and major depressive disorder.
A direction that ranks first for a skeletal diagnosis and for a urinary infection is not specific evidence for either.
A single pairing admits a clinical reading, but each reading fails to cover the rest that share the direction at the same rank.
Four families recurred, namely vascular calcification, pulmonary air-space disease, drains and stents, and biliary dilatation, and they point to age-linked cardiometabolic burden, acute illness and prior care rather than to the coded disease.
The audit also returned clean profiles, and every leading observation was direct or related evidence for bronchiectasis, pneumonia, coronary atherosclerosis and pulmonary collapse.

The alignment score ranks observation directions against a fitted probe, and Equation~\eqref{eq:logit_decomposition} shows what it leaves out.
It drops the scan-specific similarity and the scan-dependent normalization, and it replaces each observation vector by a bank-mean-centred direction whose length affects the ranking.
The observation vectors also correlate with one another, so a leading direction stands for a neighbourhood of phrases.
The score measures global semantic correspondence, leaving patient-level contribution to a separate calculation.
Turning a candidate direction into a demonstrated shortcut needs error-conditioned comparison, confirmation that the observation is present in the scan, and intervention on a known confounder.

Restricted probes matched the full-bank reference while drawing on a median bank of under half a percent of the observations.
The retained directions were the ones clinical practice associates with each syndrome, such as pleural effusion for CHF and a collapsed IVC for hypovolemia.
The semantic basis of a predictor is therefore separable from its accuracy.

Retaining the scan-specific similarity in the alignment score would turn a global direction into a patient-level contribution, and that is the next experiment.
Generative models can make a patient-specific confounder visible, but diffusion-based reconstruction has produced false structure under small perturbations, so any counterfactual needs a direct perturbation check beside it\cite{han2024instabilities}.
Grounding a larger share of the bank would raise the resolution of the reference frame.
Learned or ontology-derived rule sets would extend restriction past the phenotypes a keyword rule can reach.
Each of these steps belongs in a staged evaluation rather than in deployment, which is what early-stage clinical-evaluation and imaging reporting guidance already prescribe\cite{vasey2022decide,tejani2024claim}.
The same four-stage design should transfer to other modalities, and the observation bank would then be mined from whichever reports already exist.

Our study has limitations.
PMBB outcomes were mined from reports rather than read from images, and the chest rules were calibrated against CT-RATE silver labels, so the two share a common origin\cite{irvin2019chexpert,smit2020chexbert}.
External evidence came from one biobank, one public trauma dataset and one CTPA source, and the 221-phenotype criterion used test-set prevalence.
Beyond the Holm-adjusted RadLex analysis, we left multiplicity unadjusted, so those comparisons are descriptive.
The intervals in the annotation and audit figures reflect variation between findings or between probe fits rather than patient sampling.
Above all, the audit is observational and global, so nothing in this study shows that a confounder was used or removed in any individual patient.

\section*{Methods}

\subsection*{Paired CT scan, radiology report and linked EHR datasets}

A radiology report supplies two kinds of supervision from a single document, the impression and the findings section.
We drew on chest and abdominal CT, the two anatomies for which large paired CT and report corpora are publicly released.
This retrospective computational study was a secondary analysis of existing data, with no prospective recruitment and no intervention, and every source dataset was used under its applicable licence, research-use agreement and institutional approval (see the section `Ethics statement').
To assemble the development corpus, we combined two public collections of paired CT volumes and radiology reports: CT-RATE\cite{hamamci2025generalist}, which releases non-contrast chest CT, and the Merlin abdominal CT dataset\cite{blankemeier2024merlin}, which releases contrast-enhanced abdominal CT.
This gave 72,640 paired scan-report rows from 38,317 patients, of which 47,146 rows from 20,000 patients were contributed by CT-RATE (65\%) and 25,494 rows from 18,317 patients by Merlin (35\%).
Data collected for each patient are outlined below, and the cohorts held out for evaluation are described in the section `Native volume-report model and observation bank'.

\noindent\textbf{Radiology reports.}
We compiled the radiology report paired with each CT volume.
These reports consist of several sections, chiefly the findings and impression sections.
The findings section enumerates observations organ system by organ system.
The impression states the most clinically important of those findings in a few sentences.
We used the impression for contrastive pretraining and the findings section to build ACT's observation vocabulary.

\noindent\textbf{Linked EHRs.}
One cohort supplied linked EHR data.
INSPECT\cite{huang2023inspect} pairs each CTPA examination with the structured record of the encounter that produced it, and we used the diagnosis codes from that record.
To link diagnosis codes with scans, we collected the codes assigned during the same EHR visit as the CTPA.
We mapped the codes with PheWAS Phecode Map v1.2\cite{denny2010phewas,wei2017evaluating,wu2019mapping}, which groups clinically related codes into hierarchical phenotypes.
We then applied phenotype expansion, whereby a positive label for a specific phenotype during a visit was propagated to every less specific phenotype above it in the hierarchy.
Grouping the INSPECT codes in this way resulted in 1,692 visit-level phenotypes.
For each CTPA we therefore obtained an associated binary vector, with 1 indicating that a phenotype was recorded during the linked visit and 0 indicating that it was not.
Examinations without a linked visit were excluded rather than labelled all-negative.

\subsection*{Native volume-report model and observation bank}

\noindent\textbf{CT scan preprocessing.}
We reoriented all volumes to the closest canonical orientation with right, anterior and superior axes, then transposed them so that the axes ran depth, height and width.
We clipped Hounsfield units to $[-1{,}000,1{,}000]$ and resized by aspect-ratio-preserving trilinear interpolation to fit inside $160\times224\times224$ voxels, then padded symmetrically with $-1{,}000$.
Physical voxel spacing was left unresampled.
Images were left unaugmented.

\noindent\textbf{Radiology report preprocessing.}
We paired each scan with the English-language impression section of its report.
For the 5,262 Merlin reports released without an impression, we generated one from the findings using MedGemma-1.5-4B-IT\cite{sellergren2025medgemma}, prompted with few-shot examples that mapped findings to impressions.
Supplied and generated impressions were then processed identically.
Deterministic steps removed markup and administrative text, normalized whitespace and de-duplicated repeated sentences.
We tokenized the result with the contrastive language-image pretraining (CLIP) byte-pair encoder, truncated it to a 77-token context and terminated it with an end-of-text token.

\noindent\textbf{Observation-bank construction.}
ACT's observation vocabulary was constructed from the findings sections of the same reports.
After whitespace normalization there were 22,773 distinct findings sections among the CT-RATE training reports and 25,477 among the Merlin reports.
We gave Qwen3.5-35B-A3B\cite{qwen2026qwen35} two in-context examples and the question ``What are the descriptive observations in the report?'', and requested output as a JavaScript Object Notation (JSON) list.
After lowercasing, trimming and exact-string de-duplication, $M=376{,}194$ distinct phrases remained.
We retained negated phrases and semantically near-duplicate phrases.
We encoded each phrase with F2LLM-v2-14B\cite{zhang2026f2llm} using last-non-padding-token pooling.
The resulting 5,120-dimensional vectors were $\ell_2$-normalized and then held fixed for image projection, phenotype probing and auditing.

\methodsanchor{meth:native-model}%
\noindent\textbf{Model architecture.}
ACT uses a registered DINOv2 vision Transformer (ViT)-B/14 for the image encoder\cite{oquab2023dinov2,darcet2024vision}.
Registers are extra learned tokens that absorb the high-norm artefacts which otherwise appear in ViT feature maps.
Each axial slice was encoded independently into a 768-dimensional vector.
We appended a learnable class token to the end of the slice sequence and fused the sequence with a two-layer Transformer using 12 attention heads, feed-forward multiplier 2, pre-normalization and rotary positional embeddings.
The final class-token state was linearly projected to the 768-dimensional image embedding $\mathbf{z}_I$.
We used a randomly initialized CLIP-style causal text Transformer with width 512, 12 layers and eight attention heads, and projected its end-of-text state to the shared 768-dimensional embedding $\mathbf{z}_T$\cite{radford2021learning}.
\suppfigref{fig:arch} shows the complete architecture.

\noindent\textbf{Model training.}
We used the symmetric information noise-contrastive estimation (InfoNCE) objective of CLIP\cite{radford2021learning} between $\ell_2$-normalized volume and report embeddings, with a learned temperature and both retrieval directions.
Training batches sampled CT-RATE and Merlin in proportion to dataset size.
We optimized DINOv2, the fusion module, the report tower, the projection layers and the temperature jointly, using 16-bit floating-point mixed precision and AdamW with a learning rate of $1\times10^{-4}$ and weight decay of 0.2.
The per-device batch held four examples, and embeddings were not gathered across devices, so each contrastive pool held four candidates.
We accumulated gradients over 32 mini-batches to reach an effective update size of 256 examples.
The schedule used 500 warm-up updates followed by cosine decay, and training ran for at most 40 epochs.
Every 200 mini-batches we evaluated mean AUROC across the 18 CT-RATE validation labels and used it to select the checkpoint.

\methodsanchor{meth:concept-representation}%
\noindent\textbf{Concept-anchored CT representation.}
The native model and the observation bank live in different spaces, and ACT's concept-anchored CT representation is what joins them.
It projects the native 768-dimensional image-text space through the bank into the 5,120-dimensional F2LLM observation space, adapting CLEAR's concept-anchored projection and probe-observation audit to volumetric CT\cite{han2026clear}.
For a nonzero vector $\mathbf{v}$, let $\mathcal{N}(\mathbf{v})=\mathbf{v}/\lVert\mathbf{v}\rVert_2$.
Throughout, a hat marks a vector that has been $\ell_2$-normalized, and a tilde marks one that was centred on the observation-bank mean before being $\ell_2$-normalized.
Let $\mathbf{c}_j^0\in\mathbb{R}^{768}$ and $\mathbf{e}_j^0\in\mathbb{R}^{5{,}120}$ be the native-model and F2LLM text embeddings, respectively, of observation $j\in\{1,\ldots,M\}$.
We formed $\hat{\mathbf{c}}_j=\mathcal{N}(\mathbf{c}_j^0)$ and $\hat{\mathbf{e}}_j=\mathcal{N}(\mathbf{e}_j^0)$, and stacked their transposes as the rows of $\mathbf{C}\in\mathbb{R}^{M\times768}$ and $\mathbf{E}\in\mathbb{R}^{M\times5{,}120}$.
For a CT volume $x$, we wrote $\hat{\mathbf{z}}_I(x)=\mathcal{N}(\mathbf{z}_I(x))$ for the normalized native image embedding.
Because both vectors are $\ell_2$-normalized, $s_j(x)=\hat{\mathbf{z}}_I(x)^{\mathsf T}\hat{\mathbf{c}}_j$ is the cosine similarity between the volume and observation $j$, and $\mathbf{s}(x)=\mathbf{C}\hat{\mathbf{z}}_I(x)$ collects all $M$ of them.
The semantic image embedding is the weighted sum of the F2LLM observation embeddings, $\mathbf{e}_{\mathrm{CT}}(x)=\mathbf{E}^{\mathsf T}\mathbf{s}(x)=\sum_{j=1}^{M}s_j(x)\hat{\mathbf{e}}_j\in\mathbb{R}^{5{,}120}$.
ACT's concept-anchored CT representation is its normalization:
\begin{equation}
\hat{\mathbf{h}}(x)=\mathcal{N}\!\left(\mathbf{E}^{\mathsf T}\mathbf{C}\hat{\mathbf{z}}_I(x)\right).
\label{eq:concept_projection}
\end{equation}
In words, a volume is described by how similar it is to every observation in the bank, and those similarities then act as weights on the observations' own text embeddings.
Equation~\eqref{eq:concept_projection} uses all $M=376{,}194$ similarities.
Because $\mathbf{E}^{\mathsf T}\mathbf{C}$ is fixed once the bank is built, Equation~\eqref{eq:concept_projection} applies a single linear map of rank at most 768 to $\hat{\mathbf{z}}_I(x)$ and then normalizes.
This embedding $\hat{\mathbf{h}}(x)$ is the representation that every downstream phenotype probe acts on, and the fitted weights of those probes are what the probe-observation audit ranks against the observation directions.

\noindent\textbf{Data splits.}
We divided the development corpus so that pretraining never saw an evaluation scan.
Pretraining used the CT-RATE training split and all three released Merlin partitions, namely 15,314 training, 5,055 validation and 5,125 test scans.
Every Merlin partition therefore stayed out of evaluation.
The CT-RATE validation split was reserved for checkpoint selection alone, and the CT-RATE test split of 2,489 non-contrast chest CT scans from 973 patients, carrying 18 thoracic finding labels, became the in-domain annotation cohort.
The external annotation cohorts were PMBB\cite{verma2022penn} and RSNA-2023\cite{rsna2023abdominal}.
PMBB contributed 9,097 non-contrast chest scans carrying the same 18 thoracic findings and 14,290 contrast-enhanced abdominal scans carrying 30 abdominal findings.
Each regional PMBB pool held at most one scan per patient, and 2,029 patients occurred in both pools, giving 23,387 scans from 21,358 unique patients.
RSNA-2023 contributed 943 contrast-enhanced abdominal scans in 629 reconstructed patient clusters with nine abdominal-trauma outcomes.
CT-RATE and RSNA-2023 carry labels released with each dataset, whereas the PMBB labels were mined from its reports, as described in the section `Component evaluations'.
INSPECT supported the CTPA-EHR phenotype experiments and entered no other analysis, with 15,352 scans from 12,992 patients for training, 901 scans from 766 patients for validation and 2,612 scans from 2,223 patients for testing.
We ensured that scans from a single patient did not appear in more than one INSPECT partition, and none of the PMBB, RSNA-2023 or INSPECT volumes was used for pretraining.

\subsection*{Component evaluations}

\noindent\textbf{Zero-shot finding annotation.}
For each finding we computed the cosine similarities between the normalized image embedding and the normalized text embeddings of two prompts, the dataset-specific finding name and its ``no [finding]'' counterpart.
For pleural effusion, for example, the two prompts were ``pleural effusion'' and ``no pleural effusion''.
We converted the two similarities to a positive probability with a two-class softmax and no additional temperature\cite{tiu2022chexzero}.
\figref{fig:annotation} used ACT's native image embedding $\mathbf{z}_I$ and native text tower, not the concept-anchored representation $\hat{\mathbf{h}}(x)$ of Equation~\eqref{eq:concept_projection}.

\noindent\textbf{Finding-label provenance.}
CT-RATE evaluation used the dataset's released RadBERT-derived silver labels\cite{hamamci2025generalist}, and RSNA-2023 used its released reference outcomes\cite{rsna2023abdominal}.
PMBB releases no finding labels, so we created them by deterministic phrase mining of the findings, anatomy and impression sections, without an LLM.
We excluded the history, indication, technique, comparison and recommendation sections.
Sentence-scoped negation and uncertainty rules assigned present, uncertain or absent status.
The chest rules were calibrated against CT-RATE's RadBERT silver labels in that 2,489-scan test split, reaching a macro-F1 score of 0.879 and micro precision, recall and F1 scores of 0.94, 0.86 and 0.90.

\noindent\textbf{Baseline models.}
We compared ACT's native model with two released 3D vision-language systems, Merlin\cite{blankemeier2024merlin} and CT-CLIP\cite{hamamci2025generalist}, and three 2D systems, MedSigLIP\cite{sellergren2025medgemma}, BiomedCLIP\cite{zhang2025multimodal} and OpenAI CLIP\cite{radford2021learning}.
To keep the cohort and finding denominators fixed across models, the main benchmark included only baselines that returned row-aligned scores for every scan and every finding in a cohort.
Each model retained its own released or study-defined preprocessing, tokenizer, text tower and prompt template.
For the 2D systems we converted each volume to three-channel red, green and blue images, generating the channels with wide ($[-1{,}024,1{,}024]$), soft-tissue ($[-135,215]$) and narrow ($[0,80]$) Hounsfield unit windows.
We sampled 85 axial slices uniformly from each volume and mean pooled their embeddings.

\noindent\textbf{Finding-conditioned retrieval.}
Concept-to-image retrieval used each finding score to rank volumes.
Pooled Recall@$k$ estimated the expected recall in a 64-candidate pool holding one positive and 63 negatives sampled with replacement.
For each positive scan, if a fraction $q$ of the eligible negatives ranked before it, we computed the top-$k$ probability as $\Pr\{\operatorname{Binomial}(63,q)\leq k-1\}$.
In words, instead of drawing one pool of 63 negatives and counting hits, we computed the probability that fewer than $k$ of them would outrank the positive, which is the expected recall over all such pools and removes the sampling noise of a single draw.
We averaged those probabilities over positives and then over findings.
Random ranking gives floors of 1.6\%, 7.8\% and 15.6\% at Recall@1, @5 and @10.
For image-to-concept retrieval we standardized scores within each finding, then had each scan rank the cohort's finding vocabulary, and averaged Recall@1, @3 and @5 over scans with at least one positive finding.
The finding vocabulary and the within-finding standardization parameters were fixed before bootstrap resampling.

\noindent\textbf{Concept score versus annotated lesion volume.}
For \figref[d]{fig:annotation} we used ReXGroundingCT, which links expert 3D segmentations to individual report phrases for a subset of CT-RATE\cite{baharoon2026rexgroundingct}.
Of its 3,142 annotated scans, 332 derive from the CT-RATE validation split and lie in our 2,489-volume held-out test set, and 313 of these, from 283 patients, carried at least one finding in an evaluated category.
None was seen during pretraining.
ReXGroundingCT's exhaustively radiologist-annotated validation and test splits derive from CT-RATE's training split and were excluded, so we used its training-split annotations, segmented by trained annotators and reviewed by a board-certified radiologist.
Volumes were the segmented voxel count multiplied by the voxel volume of the paired CT, summed when a scan carried several findings of one category.
The annotations segment at most three instances per finding, so plotted volumes are lower bounds for multifocal findings.
Concept scores were computed exactly as in the zero-shot analyses.
Each category was scored with candidate prompts covering the category name, its defining sub-terms, the corresponding CT-RATE label and common synonyms, 45 prompts across the evaluated categories.

\noindent\textbf{High-score montage and slice selection.}
For the qualitative montage in \figref[e]{fig:annotation}, we subtracted a finding-free text reference from each prompt embedding to reduce prompt bias\cite{kim2024transparent}.
For each of five findings we selected ten volumes among the 30 highest-scoring held-out PMBB volumes.
We displayed each volume at the axial slice with the largest positive attribution computed from gradients and activations, and we did not overlay the attribution map.

\noindent\textbf{Observation-bank lexical summary.}
Independently of the RadLex-anchored grouping described below, we assigned each observation to one of 18 named organ-system categories or to Other/unspecified.
We counted distinct matches to a prioritized keyword lexicon, retained the category with the most matches, broke ties by a fixed category order and assigned unmatched phrases to Other/unspecified.
\suppfigref{fig:concept_wordclouds} displays the 16 largest named categories, with word size representing unigram or collocation frequency.
Before counting we removed English stop words together with study-defined unit, position and reporting-filler terms, and we normalized plurals.
We used all assigned observations except in Lung \& airways, where we drew a fixed-seed ($7$) random sample of 70,000 of 82,318 observations.

\noindent\textbf{Anatomical families and UMAP.}
We reduced the observation embeddings to 50 components by randomized principal component analysis (PCA), a space we refer to as PCA-50 and reuse for the external mapping below.
We then fitted uniform manifold approximation and projection (UMAP)\cite{mcinnes2018umap} with 15 neighbours, a minimum distance of 0.5, a spread of 1 and Euclidean distance.
The plotted coordinates comprise all 376,194 observations without subsampling.
UMAP was fitted without a fixed random seed and served only for visualization.
We coloured the points using 14 prespecified anatomical families defined by explicit anatomy and finding rules anchored to active terms in RadLex 4.3\cite{langlotz2006radlex,rubin2008terminology,rsna2025radlex}.
These groupings are project-defined rather than official RadLex classes.
We assigned unresolved anatomical sites, non-anatomical or device-related phrases and equal-specificity multisystem observations to Other.

\noindent\textbf{RadLex ontology analysis.}
We parsed direct named subclass edges from the Clinical finding branch of RadLex 4.3 (RID34785).
Observations were grounded independently of their embeddings, by conservative exact token-span matching to unambiguous active English labels, synonyms and acronyms, with ambiguous and obsolete terms excluded.
This procedure mapped 20,307 of the 376,194 observations (5.4\%) to 361 RadLex identifiers.
For shortest-path distances from one to five we sampled up to 200 identifier pairs and 100 observation pairs per identifier pair, then measured Euclidean distance in the full 5,120-dimensional space rather than in PCA-50.

\noindent\textbf{External PMBB fixed-atlas mapping.}
A separate Qwen3.6-35B-A3B extraction covered 16,896 PMBB reports, namely all 9,097 chest reports and 7,799 of the 14,290 abdominal reports.
The remaining 6,491 abdominal reports stayed outside this extraction.
Extraction yielded 390,111 observation occurrences and 157,466 unique lowercased and exactly de-duplicated phrases, which we embedded with the same F2LLM model as the reference bank.
We placed these phrases in the unchanged reference atlas by 15-neighbour, UMAP-membership-weighted barycentric reconstruction, after exact neighbour reranking in reference PCA-50 space.
Neither the reference PCA nor the UMAP coordinates were refitted.
Exact normalized strings already present in the reference bank retained their reference coordinates but were excluded from the PMBB-only overlays and from source-specific centroids.
For the source hierarchy we crossed the anatomical assignments with 18 prespecified finding families and retained intersections holding at least 100 reference and 20 PMBB-only observations.
Family centroids were computed in reference PCA-50 space, compared using Pearson distance and joined by complete linkage.

\subsection*{CTPA-EHR phenotype experiments}

The remaining experiments all sit on INSPECT and all fit linear probes, while the native image encoder and the observation bank stay frozen.
We first asked whether ACT's concept-anchored CT representation carries phenotype information in an acquisition domain absent from image pretraining.
We then asked which named observation directions the fitted probes align with, and finally whether restricting the bank to phenotype-specific observations redirects a probe without costing held-out classification.

\noindent\textbf{Phenotype label construction.}
The phecode labels are those built in the section `Paired CT scan, radiology report and linked EHR datasets'.
From the 1,692 visit-level phenotypes we retained the 221 with at least 50 positive scans in the held-out test split.
This eligibility criterion is post hoc and uses test-label prevalence.
It selects which phenotypes are reported, and the test labels themselves stayed outside both probe fitting and validation AUROC, but the resulting phenotype set was still in place during probe fitting and learning-rate selection.
The phecodes remain encounter-derived weak labels rather than adjudicated CTPA findings.

\noindent\textbf{Phenotype scoring and linear probing.}
Zero-shot scores used the same two-class softmax procedure as finding annotation, described in the section `Component evaluations'.
For ACT's concept-anchored CT representation the positive and negative strings were the bare phenotype name and ``no [phenotype]''.
For CT-CLIP they were ``[phenotype] is present.'' and ``[phenotype] is not present.'', which is the prompt form that model was released with.
For supervised evaluation every image encoder and representation transformation remained frozen, and we optimized a multilabel linear sigmoid head with mean binary cross-entropy and no class weighting.
\figref{fig:phenotyping} used the same patient-disjoint partitions for both representations, ACT's 5,120-dimensional concept-anchored embedding and CT-CLIP's native 512-dimensional embedding.
Probe optimization used Adam for at most 200 epochs, mini-batches of 512, weight decay $10^{-8}$ and validation macro-AUROC for early stopping with patience 10.
Learning rates were selected on the validation split for each model separately, giving 0.03 for ACT's representation and 0.005 for CT-CLIP.
Each fit was repeated 20 times from different initializations, and the validation-selected probe state was evaluated on the test split.

\methodsanchor{meth:probe-audit}%
\noindent\textbf{Probe-observation audit.}
The audit reuses the 20 probe fits just described and asks which observation directions each fitted probe points along.
Let $\mathbf{w}_q^{(r)}\in\mathbb{R}^{5{,}120}$ denote the weight vector of the probe for phenotype $q$ in fit $r$, with $r\in\{1,\ldots,R\}$ and $R=20$.
We centred each observation on the bank mean $\bar{\mathbf{e}}=M^{-1}\sum_{j=1}^{M}\hat{\mathbf{e}}_j$ and renormalized, giving the direction $\tilde{\mathbf{e}}_j=\mathcal{N}(\hat{\mathbf{e}}_j-\bar{\mathbf{e}})$.
A single fit gives the alignment $a_{qj}^{(r)}=(\mathbf{w}_q^{(r)})^{\mathsf T}\tilde{\mathbf{e}}_j$.
The reported score averages those over the $R$ fits:
\begin{equation}
\bar a_{qj}=\frac{1}{R}\sum_{r=1}^{R}(\mathbf{w}_q^{(r)})^{\mathsf T}\tilde{\mathbf{e}}_j.
\label{eq:audit_score}
\end{equation}
In words, we measured how strongly a probe's weight vector points along each observation's direction, then averaged that over the 20 fits.
We ranked observations separately for each phenotype by $\bar a_{qj}$, without semantic near-duplicate filtering.
Probe weights remained unnormalized, so score magnitudes are comparable within a phenotype and incomparable across phenotypes.

Each probe acts on the representation $\hat{\mathbf{h}}(x)$ of Equation~\eqref{eq:concept_projection}, so its logit decomposes as
\begin{equation}
(\mathbf{w}_q^{(r)})^{\mathsf T}\hat{\mathbf{h}}(x)+b_q^{(r)}
=\frac{1}{\lVert\mathbf{e}_{\mathrm{CT}}(x)\rVert_2}\sum_{j=1}^{M}s_j(x)\,(\mathbf{w}_q^{(r)})^{\mathsf T}\hat{\mathbf{e}}_j+b_q^{(r)},
\label{eq:logit_decomposition}
\end{equation}
where $b_q^{(r)}$ is the fitted intercept.
Writing $\hat{\mathbf{e}}_j=\bar{\mathbf{e}}+\rho_j\tilde{\mathbf{e}}_j$ with $\rho_j=\lVert\hat{\mathbf{e}}_j-\bar{\mathbf{e}}\rVert_2$ shows that observation $j$ contributes $s_j(x)\{(\mathbf{w}_q^{(r)})^{\mathsf T}\bar{\mathbf{e}}+\rho_j a_{qj}^{(r)}\}/\lVert\mathbf{e}_{\mathrm{CT}}(x)\rVert_2$ to that logit.
The audit score differs from this contribution in three ways.
It omits the scan-specific similarity $s_j(x)$, it drops the scan-dependent denominator $\lVert\mathbf{e}_{\mathrm{CT}}(x)\rVert_2$, and it replaces $\hat{\mathbf{e}}_j$ by the centred direction $\tilde{\mathbf{e}}_j$, whose observation-dependent scale $\rho_j$ affects the ranking.
The score therefore measures global semantic alignment between a probe and an observation direction, not the contribution of an observation to an individual prediction.

\methodsanchor{meth:clinical-screen}%
\noindent\textbf{Rule-based clinical screen.}
Throughout, phenotype-related means clinically related to the coded phenotype without defining it.
A clinician-defined keyword and regular-expression screen classified phrases as direct, imaging-support, composite-branch or association-only evidence.
Explicit exclusions covered uncertainty, temporal comparison, procedure or device evidence, target mismatch and insufficient target specificity.
The procedural exclusion matched singular, plural and derived forms of device and procedure terms, including drains, drainage, catheters, stents, tubes, sutures, staples, mesh and anastomotic references.
Terms that name anatomy or pathology rather than a device were deliberately retained, namely rectus and peribronchial sheath, peritoneal implants of carcinomatosis, intrahepatic vascular shunts, a spontaneously draining wound or sinus tract, and calcified atheroma plates.
We applied the screen to each phenotype's natural top 10 after ranking, so it left scores and ordering unchanged.
A separate clinician-defined display-relevance review classified the observations of the profiles selected for \suppfigref{fig:clinical_alignment_audit}, and is described in Supplementary Section~A.3.

\methodsanchor{meth:bank-restriction}%
\noindent\textbf{Observation-bank restriction.}
Clinician-defined keyword and regular-expression rules, with explicit exclusions and precedence, selected direct or associated observations for each phenotype.
Among the 221 phenotypes, 54 had at least one observation naming evidence that defines the coded phenotype.
A further 32 had no such direct candidate but had at least one observation naming a clinically associated finding.
For the remaining 135, no observation matched either rule.
We retained these 86 non-empty rule sets and excluded the 135 empty ones. 
For phenotype $q$, let $J_q\subseteq\{1,\ldots,M\}$ be its non-empty index set, and let $\mathbf{C}_{J_q}$ and $\mathbf{E}_{J_q}$ denote the corresponding rows of the two observation-embedding matrices.
We then recomputed
\begin{equation}
\hat{\mathbf{h}}_{J_q}(x)=\mathcal{N}\!\left(\mathbf{E}_{J_q}^{\mathsf T}\mathbf{C}_{J_q}\hat{\mathbf{z}}_I(x)\right).
\label{eq:restricted_embedding}
\end{equation}
Equation~\eqref{eq:restricted_embedding} restricts the bank before semantic aggregation and before scan-wise normalization, rather than masking coordinates of an embedding that has already been aggregated.
The image encoder and both observation-embedding matrices remained frozen.

\noindent\textbf{Restricted probe fitting.}
We fitted one probe per phenotype to its restricted representation.
For each retained observation, the weight displayed in \figref{fig:clinical_prior_refinement} is the dot product between its normalized F2LLM vector and the fitted coefficient vector.
Selection rules for the post-hoc paired audits of \suppfigref{fig:refinement_higher_auroc_profiles} and \suppfigref{fig:refinement_lower_auroc_profiles} are given in Supplementary Section~A.3.

\subsection*{Statistical analysis}

We computed AUROC separately for each finding or phenotype.
\figref[b]{fig:annotation} reports the mean and 1.96 times the s.e.m.\ across finding-level AUROCs, so those bars reflect between-finding variation rather than patient sampling.
Percentile 95\% CIs for zero-shot annotation and for retrieval came from 1,000 patient-clustered bootstrap resamples that retained every scan from each sampled patient.
We shared those resamples across models to obtain paired model differences, and treated a difference as supported when its 95\% CI excluded zero.
Multiplicity was left unadjusted.
In the RadLex analysis, Spearman correlation quantified association with ontology path length, and adjacent distances were compared using one-sided Welch tests on identifier-pair means with Holm adjustment across four comparisons.
The linear-probe estimates in \figref{fig:phenotyping} and the probe-observation alignment summaries rest on 20 probe fits on fixed data, so their s.d.\ and two-sided $t$-based CIs quantify variation across fits rather than patient sampling.
For the overall comparison in \figref{fig:phenotyping}, percentile 95\% CIs came from 1,000 shared patient-clustered bootstrap resamples of the held-out scans, with the fitted probes and zero-shot scores fixed within resamples and paired differences taken on the shared resamples.
For the display markers in \figref{fig:clinical_prior_refinement} and in the paired supplementary audits, restricted-probe test AUROC and observation weights were summarized across 20 test-set and 20 training-set resamples, respectively.
We withheld hypothesis testing from the restriction analysis at both the per-phenotype and the aggregate level.
All remaining comparisons were descriptive and exploratory.
Supplementary Section~A.2 gives the full resampling procedures behind the supplementary AUROC tables.

\subsection*{Ethics statement}

CT-RATE was anonymized before release, and its source study received institutional ethics approval.
The retrospective Merlin and INSPECT source studies were approved by the Stanford University Institutional Review Board with waivers of informed consent, and their released data were de-identified.
PMBB participants provided informed consent for biobank participation and for research use of linked clinical and imaging data, under an Institutional Review Board-approved University of Pennsylvania biobank protocol.
PMBB data were controlled access and governed by PMBB procedures.
RSNA-2023 was a publicly released research dataset.
All datasets were used subject to their applicable licences, research-use agreements and institutional data-governance requirements.
Controlled-access data were analysed solely for research and only on institutionally managed secure computing systems.
Access was restricted to authorized study personnel trained in handling sensitive health data.
Model output stayed out of clinical care.

\section*{Data availability}

CT-RATE is available from Hugging Face (\url{https://huggingface.co/datasets/ibrahimhamamci/CT-RATE}), subject to the repository's access terms.
The Merlin abdominal CT dataset is available through AIMI (\url{https://stanfordaimi.azurewebsites.net/datasets/60b9c7ff-877b-48ce-96c3-0194c8205c40}).
Access requires completion of the dataset's data-use agreement.
RSNA-2023 is available to registered users through the corresponding abdominal trauma detection competition (\url{https://www.kaggle.com/competitions/rsna-2023-abdominal-trauma-detection}), subject to the competition rules.
INSPECT imaging and report data are available through AIMI (\url{https://stanfordaimi.azurewebsites.net/datasets?term=INSPECT}), and the linked EHR release is available through Redivis (\url{https://stanford.redivis.com/datasets/dzc6-9jyt6gapt}).
These data are provided for non-commercial research under a research-use agreement.
Individual-level PMBB images, reports, report-derived labels and the full PMBB-derived observation bank remain restricted because they are pseudonymized participant data.
Eight illustrative observation strings are reported in Supplementary Table~\ref{tab:pmbb-polarity-neighbors}.
Access is governed by PMBB and requires an eligible Penn collaboration, applicable institutional approval and a data-access agreement.
Information for researchers is available at \url{https://pmbb.med.upenn.edu/investigators.php}.
Aggregate numerical data underlying the figures will be provided as Source Data with the published article.

\section*{Code availability}

Code for ACT is publicly available at \url{https://github.com/peterhan91/ACT}.
The release covers native CT preprocessing, volume-report model training and inference, observation-bank extraction and embedding, construction of the concept-anchored CT representation, phenotype probing, probe-observation audit, observation-bank restriction, statistical analysis and figure generation.
It also provides a versioned software environment, prompt files and phenotype-rule manifests.
Weights for ACT's native volume-report model are available for academic research at \url{https://huggingface.co/peterhan91/clip_3d_ct}.
The redistributable observation-bank artefacts derived from CT-RATE and Merlin will be released at the same address.
The source datasets' licences and data-use terms apply.
We will not release patient-level PMBB images, reports, identifiers, report-derived labels or the full PMBB-derived observation bank, nor row-level INSPECT data or labels.

\section*{Author contributions}

R.W., D.T. and T.H. conceived the study and designed the ACT framework.
R.W. and T.H. developed ACT's native 3D volume-report model and software, conducted its image-text pretraining and performed the zero-shot and supervised evaluation experiments.
R.W. and T.H. also constructed the report-derived observation bank, implemented ACT's concept-anchored CT representation and performed the semantic-organization, phenotype-audit and observation-bank-restriction analyses.
W.R.W. and E.M.B. supported access to and interpretation of the PMBB data.
Y.L., F.B.O., K.K.B., L.C.A., G.E.W., L.S., C.D., E.M.B. and D.T. provided clinical and methodological input and interpreted the results.
R.W., D.T. and T.H. analysed the results and prepared the figures.
D.T. and T.H. jointly supervised the study.
R.W., D.T. and T.H. wrote the manuscript with input from all authors.
All authors reviewed and revised the manuscript and approved the final version.

\section*{Competing interests}

K.K.B. reports speaker fees from Canon Medical Systems and GE Healthcare and participation on a data safety monitoring board for Philips (IHI Project IMAGIO).
D.T. holds shares in StratifAI and Synagen and has received honoraria from AstraZeneca, MSD, Roche, Siemens and Philips.
The remaining authors declare no competing interests.

\clearpage
\bibliographystyle{unsrt}

\clearpage
\begingroup
\setcounter{page}{1}
\normalsize
\setstretch{1.0}
\setlength{\parskip}{0.75em}
\setlength{\parindent}{0pt}

\renewcommand{\figurename}{Supplementary Fig.}
\renewcommand{\tablename}{Supplementary Table}
\captionsetup[table]{labelfont=bf}
\renewcommand{\thefigure}{\arabic{figure}}
\renewcommand{\thetable}{\arabic{table}}
\setcounter{figure}{0}
\setcounter{table}{0}
\setcounter{section}{0}
\renewcommand{\thesection}{\Alph{section}}
\renewcommand{\thesubsection}{\thesection.\arabic{subsection}}
\setcounter{tocdepth}{2}
\titlespacing*{\subsection}{0pt}{1.5em}{0.45em}

\noindent{\Large\bfseries ACT Supplementary Information\par}
\vspace{1.6em}

\makeatletter
\@starttoc{toc}
\makeatother
\clearpage

\section{Extended Methods}

\subsection{Architecture of ACT's native 3D volume-report model}

\begin{figure}[H]
    \centering
    \includegraphics[width=0.8\textwidth]{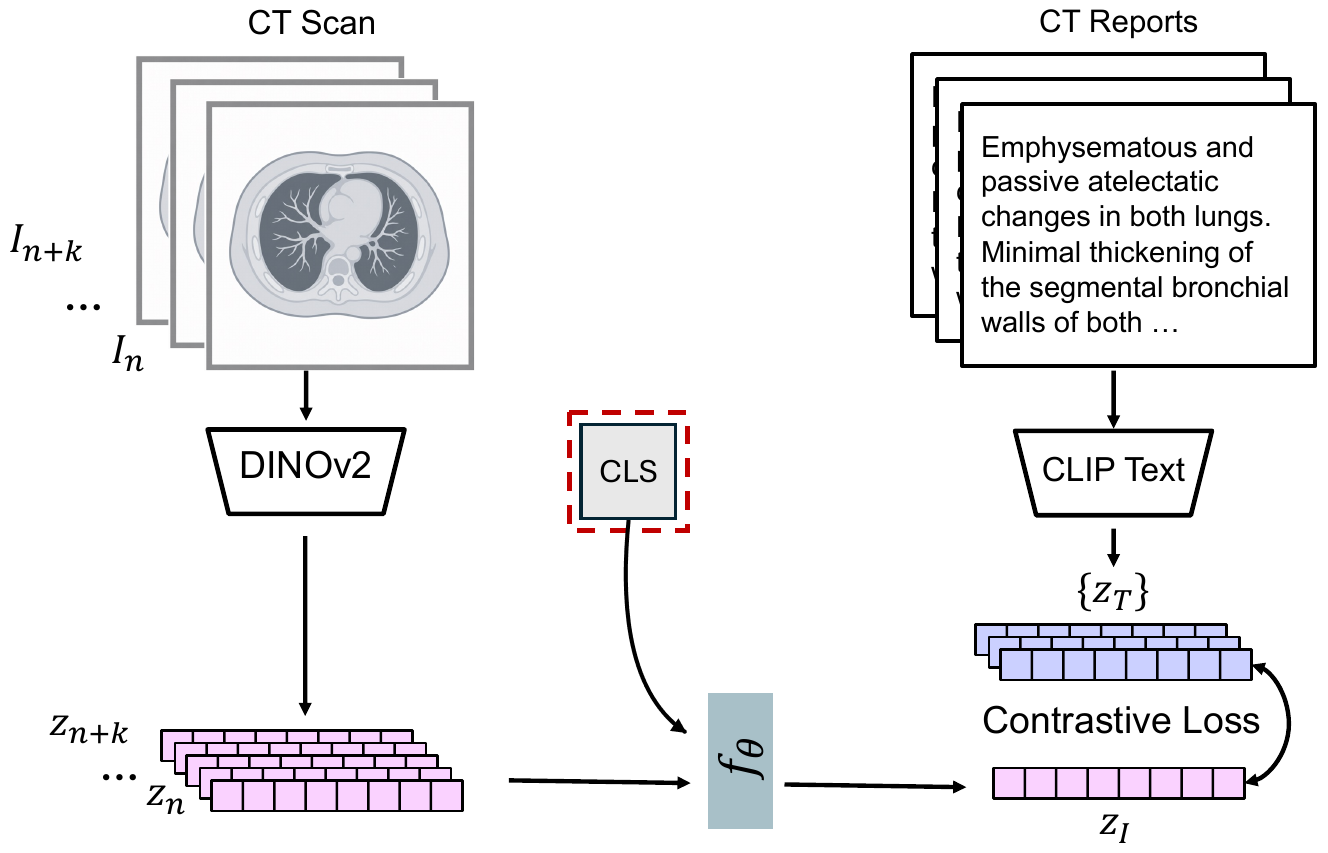}
    \caption{\textbf{Contrastive pretraining of ACT's native 3D volume-report model.}
    Each CT volume is resized and padded to $160\times224\times224$ (slices $\times$ height $\times$ width) and represented as a sequence of axial slices.
    A DINOv2 ViT-B image encoder embeds each slice independently.
    The Transformer aggregator $f_\theta$ fuses a learnable classification (CLS) token and the slice embeddings into the volume embedding $\mathbf{z}_I$.
    In parallel, a CLIP-style causal text Transformer encodes the paired radiology impression into $\mathbf{z}_T$.
    A symmetric contrastive objective aligns $\mathbf{z}_I$ and $\mathbf{z}_T$ during pretraining on 72,640 paired scan-report rows from 38,317 patients.}
    \label{fig:arch}
\end{figure}

\subsection{Uncertainty estimation for supplementary AUROC tables}

\noindent\textbf{Zero-shot finding AUROCs.}
The point estimate was the empirical full-cohort AUROC for every eligible combination of cohort, finding and model with complete row-aligned predictions.
We estimated 95\% confidence intervals (CIs) using 1,000 patient-clustered bootstrap resamples, sampling patients with replacement and assigning their multiplicity to all associated scans.
The same draws were reused across eligible models and findings within each cohort, and the 2.5th and 97.5th percentiles of valid resampled AUROCs defined the CI.
Each PMBB regional cohort contained one selected scan per patient, making patient- and scan-level resampling equivalent within that cohort.
These CIs condition on fixed predictions and characterize patient-sampling variability in the test cohort, unlike the between-finding error bars in \figref[b]{fig:annotation}.

\begin{samepage}
For RSNA-2023, neither the original splitter nor its emitted split manifest was retained.
We therefore reconstructed the recorded 70/10/20 seed-42 patient split and the lexicographic series ordering from the retained run script, contemporaneous pipeline conventions and the released series metadata\cite{rsna2023abdominal}.
This yielded 943 series in 629 reconstructed patient clusters.
We paired these clusters with the exact retained $943\times9$ outcome matrix used for evaluation rather than the earlier competition label version.
Before resampling, we required a constant outcome vector within each reconstructed patient.
We also required the reconstruction to reproduce all 54 displayed model-finding point AUROCs.
These checks establish consistency with the retained evaluation row order but stop short of independently recovering patient identity.
\end{samepage}

\noindent\textbf{CTPA phenotype-probe AUROCs.}
For each phenotype and representation, the point estimate was the mean test AUROC over the same 20 matched probe fits reported in \figref{fig:phenotyping}.
We report a two-sided Student $t$ CI across those fits.
Fit by fit, we calculated paired differences between ACT and CT-CLIP, then constructed their intervals.
These CIs characterize fitted-probe variation on a fixed test set.
For the overall comparison, we used 1,000 shared patient-clustered bootstrap resamples of the 2,612 held-out scans from 2,223 patients, drawn as in the restriction analysis below.
Within each resample, we recomputed macro AUROC across the 221 phenotypes for each fixed model, averaging the 20 matched probe fits in the linear-probe regime.
Paired ACT minus CT-CLIP macro differences used the shared resamples, and the 2.5th and 97.5th percentiles defined the reported 95\% CIs.

\noindent\textbf{Observation-bank restriction AUROCs.}
For Supplementary Table~\ref{tab:supp-restriction-all86}, we used 1,000 shared patient-clustered bootstrap resamples of the 2,612 held-out scans from 2,223 patients.
Within each resample, we recomputed one AUROC per probe for the fixed full-bank and rule-restricted probes.
The 2.5th and 97.5th percentiles defined the reported 95\% CIs.
Models stayed fixed within resamples, so both columns quantify held-out patient-sampling variability.
We skipped phenotype-level and aggregate hypothesis tests.

\subsection{Selection rules for supplementary audit displays}

\noindent\textbf{Shared-direction panels.}
\suppfigref{fig:shortcut_audit} was populated by shared leading direction rather than by interpretation category.
We counted how many of the 221 phenotypes carried each distinct rank-1 observation string.
The most widely shared string covered 20 phenotypes and the next covered 18, but both describe atherosclerotic vascular calcification, so the second panel instead used the most widely shared string naming a different finding, biliary ductal dilatation, which covered 15 phenotypes.
Every phenotype carrying a selected string at rank 1 is displayed, 20 for the first and 15 for the second, so phenotype-level selection stayed out of the panels.
Selection used rank-1 identity alone, with interpretation category, AUROC threshold and clinical filter all withheld.
Each cell holds the observation occupying that natural rank for that phenotype, and its fill gives the percentage of the phenotypes in the panel whose own mean-ranked top 5 contains that observation.
Alignment magnitudes are omitted, so the display never compares score magnitudes between phenotypes.
Selection and category assignment had no independent clinical adjudication, and they illustrate candidate proxy hypotheses rather than estimate how often such directions occur.
We retained the complete $221\times25$ ranked export, the panel-selection record and the source rows as machine-readable data.

\noindent\textbf{Display-relevance review.}
A separate clinician-defined display-relevance review classified each observation in the selected profiles as direct target evidence or as clinically related but non-defining or qualified context.
This display axis left the conservative screen intact.
An uncertain or temporal phrase could remain strictly rejected and still be shown in light blue if its finding was clinically related to the phenotype.
For \suppfigref{fig:clinical_alignment_audit} we selected four phenotypes for which all 10 natural observations met one of these two display classes, drawn from distinct three-digit phecode groups, none of which appears in \suppfigref{fig:shortcut_audit}.
That figure shows unfiltered natural mean ranks 1 to 5 from the same complete bank used in the main audit, with dark blue denoting direct target evidence and light blue denoting related context.
Neither the rule-based screen nor the display-relevance review was independently adjudicated by a radiologist.

\noindent\textbf{Paired full-bank and refined audits.}
For the post-hoc paired audits, the full-bank side shows natural mean ranks 1 to 5, and the refined side shows the five observations with the largest bootstrap-mean weights among the restricted probe's 20 largest positive weights.
After manual clinical review we chose four higher-AUROC and four lower-AUROC examples from the 86 eligible phenotypes, excluding the eight already shown in \figref{fig:clinical_prior_refinement}.
Both displayed top-five profiles kept their raw form, free of near-duplicate removal, uncertainty filtering and manual reranking.

\section{Extended Results}

\subsection{Native-model annotation and retrieval}

\begingroup
\normalsize
\setlength{\tabcolsep}{2pt}
\renewcommand{\arraystretch}{1.0}
\setlength{\LTcapwidth}{\linewidth}
\setlength{\LTleft}{0pt}
\setlength{\LTright}{0pt plus 1fill}
\begin{longtable}{@{}>{\raggedright\arraybackslash}p{0.260\linewidth}>{\centering\arraybackslash}p{0.060\linewidth}>{\centering\arraybackslash}p{0.210\linewidth}>{\centering\arraybackslash}p{0.210\linewidth}>{\centering\arraybackslash}p{0.210\linewidth}@{}}
\caption{\textbf{Zero-shot finding annotation on CT-RATE test (model panel 1 of 2).} Values are AUROC (95\% percentile CI) from 1,000 shared patient-cluster bootstrap resamples. Bold indicates the largest point estimate across all model panels for each finding. $n_{+}$ is the number of positive test scans.}\label{tab:supp-zs-ctrate-test-a}\\
\toprule
\textbf{Finding} & \textbf{$n_{+}$} & \textbf{ACT} & \textbf{Merlin} & \textbf{CT-CLIP} \\
\midrule
\endfirsthead
\multicolumn{5}{c}{\tablename\ \thetable\ -- continued}\\
\toprule
\textbf{Finding} & \textbf{$n_{+}$} & \textbf{ACT} & \textbf{Merlin} & \textbf{CT-CLIP} \\
\midrule
\endhead
\midrule
\multicolumn{5}{r}{Continued on next page}\\
\endfoot
\bottomrule
\endlastfoot
Hiatal hernia & 341 & \mbox{0.580 (0.535, 0.623)} & \mbox{0.561 (0.517, 0.608)} & \textbf{\mbox{0.706 (0.662, 0.745)}} \\
Medical material & 277 & \textbf{\mbox{0.718 (0.668, 0.767)}} & \mbox{0.548 (0.502, 0.597)} & \mbox{0.704 (0.656, 0.753)} \\
Lymphadenopathy & 693 & \textbf{\mbox{0.717 (0.683, 0.749)}} & \mbox{0.595 (0.559, 0.625)} & \mbox{0.704 (0.668, 0.735)} \\
Lung nodule & 1123 & \mbox{0.549 (0.511, 0.585)} & \mbox{0.496 (0.471, 0.521)} & \textbf{\mbox{0.584 (0.549, 0.619)}} \\
Pulmonary fibrotic sequela & 673 & \textbf{\mbox{0.592 (0.554, 0.627)}} & \mbox{0.532 (0.494, 0.568)} & \mbox{0.564 (0.523, 0.603)} \\
Lung opacity & 971 & \textbf{\mbox{0.695 (0.664, 0.726)}} & \mbox{0.664 (0.629, 0.695)} & \mbox{0.627 (0.592, 0.659)} \\
Atelectasis & 600 & \textbf{\mbox{0.716 (0.681, 0.750)}} & \mbox{0.598 (0.561, 0.634)} & \mbox{0.684 (0.648, 0.718)} \\
Emphysema & 496 & \textbf{\mbox{0.729 (0.693, 0.763)}} & \mbox{0.604 (0.563, 0.644)} & \mbox{0.727 (0.691, 0.765)} \\
\end{longtable}
\endgroup
\begingroup
\normalsize
\setlength{\tabcolsep}{2pt}
\renewcommand{\arraystretch}{1.0}
\setlength{\LTcapwidth}{\linewidth}
\setlength{\LTleft}{0pt}
\setlength{\LTright}{0pt plus 1fill}
\begin{longtable}{@{}>{\raggedright\arraybackslash}p{0.260\linewidth}>{\centering\arraybackslash}p{0.060\linewidth}>{\centering\arraybackslash}p{0.210\linewidth}>{\centering\arraybackslash}p{0.210\linewidth}>{\centering\arraybackslash}p{0.210\linewidth}@{}}
\caption{\textbf{Zero-shot finding annotation on CT-RATE test (model panel 2 of 2).} Values are AUROC (95\% percentile CI) from 1,000 shared patient-cluster bootstrap resamples. Bold indicates the largest point estimate across all model panels for each finding. $n_{+}$ is the number of positive test scans.}\label{tab:supp-zs-ctrate-test-b}\\
\toprule
\textbf{Finding} & \textbf{$n_{+}$} & \textbf{MedSigLIP} & \textbf{BiomedCLIP} & \textbf{OpenAI CLIP} \\
\midrule
\endfirsthead
\multicolumn{5}{c}{\tablename\ \thetable\ -- continued}\\
\toprule
\textbf{Finding} & \textbf{$n_{+}$} & \textbf{MedSigLIP} & \textbf{BiomedCLIP} & \textbf{OpenAI CLIP} \\
\midrule
\endhead
\midrule
\multicolumn{5}{r}{Continued on next page}\\
\endfoot
\bottomrule
\endlastfoot
Hiatal hernia & 341 & \mbox{0.596 (0.552, 0.640)} & \mbox{0.447 (0.402, 0.490)} & \mbox{0.556 (0.513, 0.600)} \\
Medical material & 277 & \mbox{0.521 (0.485, 0.555)} & \mbox{0.629 (0.582, 0.678)} & \mbox{0.386 (0.338, 0.434)} \\
Lymphadenopathy & 693 & \mbox{0.624 (0.600, 0.648)} & \mbox{0.492 (0.455, 0.530)} & \mbox{0.528 (0.498, 0.560)} \\
Lung nodule & 1123 & \mbox{0.506 (0.472, 0.543)} & \mbox{0.472 (0.442, 0.502)} & \mbox{0.511 (0.480, 0.543)} \\
Pulmonary fibrotic sequela & 673 & \mbox{0.501 (0.465, 0.535)} & \mbox{0.526 (0.497, 0.557)} & \mbox{0.554 (0.517, 0.592)} \\
Lung opacity & 971 & \mbox{0.671 (0.638, 0.705)} & \mbox{0.592 (0.560, 0.622)} & \mbox{0.458 (0.427, 0.489)} \\
Atelectasis & 600 & \mbox{0.675 (0.641, 0.707)} & \mbox{0.499 (0.459, 0.538)} & \mbox{0.448 (0.414, 0.486)} \\
Emphysema & 496 & \mbox{0.602 (0.562, 0.641)} & \mbox{0.523 (0.477, 0.565)} & \mbox{0.514 (0.479, 0.548)} \\
\end{longtable}
\endgroup
\begingroup
\normalsize
\setlength{\tabcolsep}{2pt}
\renewcommand{\arraystretch}{1.0}
\setlength{\LTcapwidth}{\linewidth}
\setlength{\LTleft}{0pt}
\setlength{\LTright}{0pt plus 1fill}
\begin{longtable}{@{}>{\raggedright\arraybackslash}p{0.260\linewidth}>{\centering\arraybackslash}p{0.060\linewidth}>{\centering\arraybackslash}p{0.210\linewidth}>{\centering\arraybackslash}p{0.210\linewidth}>{\centering\arraybackslash}p{0.210\linewidth}@{}}
\caption{\textbf{Zero-shot finding annotation on PMBB chest non-contrast (model panel 1 of 2).} Values are AUROC (95\% percentile CI) from 1,000 shared patient-cluster bootstrap resamples. Bold indicates the largest point estimate across all model panels for each finding. $n_{+}$ is the number of positive test scans.}\label{tab:supp-zs-pmbb-chest-nc-a}\\
\toprule
\textbf{Finding} & \textbf{$n_{+}$} & \textbf{ACT} & \textbf{Merlin} & \textbf{CT-CLIP} \\
\midrule
\endfirsthead
\multicolumn{5}{c}{\tablename\ \thetable\ -- continued}\\
\toprule
\textbf{Finding} & \textbf{$n_{+}$} & \textbf{ACT} & \textbf{Merlin} & \textbf{CT-CLIP} \\
\midrule
\endhead
\midrule
\multicolumn{5}{r}{Continued on next page}\\
\endfoot
\bottomrule
\endlastfoot
Hiatal hernia & 1214 & \mbox{0.505 (0.487, 0.521)} & \mbox{0.523 (0.508, 0.542)} & \textbf{\mbox{0.569 (0.552, 0.586)}} \\
Medical material & 2420 & \textbf{\mbox{0.705 (0.693, 0.718)}} & \mbox{0.501 (0.487, 0.514)} & \mbox{0.587 (0.575, 0.599)} \\
Lymphadenopathy & 2362 & \textbf{\mbox{0.602 (0.587, 0.616)}} & \mbox{0.527 (0.513, 0.542)} & \mbox{0.550 (0.537, 0.563)} \\
Lung nodule & 6201 & \mbox{0.493 (0.480, 0.506)} & \mbox{0.535 (0.522, 0.547)} & \mbox{0.548 (0.536, 0.561)} \\
Pulmonary fibrotic sequela & 2554 & \textbf{\mbox{0.667 (0.655, 0.679)}} & \mbox{0.585 (0.571, 0.597)} & \mbox{0.530 (0.517, 0.542)} \\
Lung opacity & 1681 & \textbf{\mbox{0.635 (0.622, 0.649)}} & \mbox{0.557 (0.542, 0.573)} & \mbox{0.529 (0.513, 0.543)} \\
Atelectasis & 3163 & \textbf{\mbox{0.673 (0.662, 0.684)}} & \mbox{0.618 (0.606, 0.631)} & \mbox{0.544 (0.532, 0.556)} \\
Emphysema & 1996 & \textbf{\mbox{0.667 (0.654, 0.679)}} & \mbox{0.600 (0.586, 0.614)} & \mbox{0.608 (0.594, 0.623)} \\
\end{longtable}
\endgroup
\begingroup
\normalsize
\setlength{\tabcolsep}{2pt}
\renewcommand{\arraystretch}{1.0}
\setlength{\LTcapwidth}{\linewidth}
\setlength{\LTleft}{0pt}
\setlength{\LTright}{0pt plus 1fill}
\begin{longtable}{@{}>{\raggedright\arraybackslash}p{0.260\linewidth}>{\centering\arraybackslash}p{0.060\linewidth}>{\centering\arraybackslash}p{0.210\linewidth}>{\centering\arraybackslash}p{0.210\linewidth}>{\centering\arraybackslash}p{0.210\linewidth}@{}}
\caption{\textbf{Zero-shot finding annotation on PMBB chest non-contrast (model panel 2 of 2).} Values are AUROC (95\% percentile CI) from 1,000 shared patient-cluster bootstrap resamples. Bold indicates the largest point estimate across all model panels for each finding. $n_{+}$ is the number of positive test scans.}\label{tab:supp-zs-pmbb-chest-nc-b}\\
\toprule
\textbf{Finding} & \textbf{$n_{+}$} & \textbf{MedSigLIP} & \textbf{BiomedCLIP} & \textbf{OpenAI CLIP} \\
\midrule
\endfirsthead
\multicolumn{5}{c}{\tablename\ \thetable\ -- continued}\\
\toprule
\textbf{Finding} & \textbf{$n_{+}$} & \textbf{MedSigLIP} & \textbf{BiomedCLIP} & \textbf{OpenAI CLIP} \\
\midrule
\endhead
\midrule
\multicolumn{5}{r}{Continued on next page}\\
\endfoot
\bottomrule
\endlastfoot
Hiatal hernia & 1214 & \mbox{0.547 (0.531, 0.563)} & \mbox{0.478 (0.462, 0.495)} & \mbox{0.494 (0.477, 0.511)} \\
Medical material & 2420 & \mbox{0.564 (0.550, 0.579)} & \mbox{0.556 (0.542, 0.569)} & \mbox{0.422 (0.408, 0.435)} \\
Lymphadenopathy & 2362 & \mbox{0.579 (0.566, 0.592)} & \mbox{0.449 (0.435, 0.463)} & \mbox{0.477 (0.463, 0.491)} \\
Lung nodule & 6201 & \mbox{0.436 (0.422, 0.449)} & \mbox{0.470 (0.458, 0.483)} & \textbf{\mbox{0.596 (0.584, 0.609)}} \\
Pulmonary fibrotic sequela & 2554 & \mbox{0.584 (0.571, 0.596)} & \mbox{0.552 (0.538, 0.564)} & \mbox{0.505 (0.492, 0.518)} \\
Lung opacity & 1681 & \mbox{0.581 (0.566, 0.596)} & \mbox{0.531 (0.515, 0.546)} & \mbox{0.479 (0.464, 0.493)} \\
Atelectasis & 3163 & \mbox{0.657 (0.645, 0.668)} & \mbox{0.530 (0.518, 0.542)} & \mbox{0.470 (0.458, 0.482)} \\
Emphysema & 1996 & \mbox{0.565 (0.551, 0.578)} & \mbox{0.516 (0.502, 0.531)} & \mbox{0.528 (0.514, 0.541)} \\
\end{longtable}
\endgroup
\begingroup
\normalsize
\setlength{\tabcolsep}{2pt}
\renewcommand{\arraystretch}{1.0}
\setlength{\LTcapwidth}{\linewidth}
\setlength{\LTleft}{0pt}
\setlength{\LTright}{0pt plus 1fill}
\begin{longtable}{@{}>{\raggedright\arraybackslash}p{0.260\linewidth}>{\centering\arraybackslash}p{0.060\linewidth}>{\centering\arraybackslash}p{0.210\linewidth}>{\centering\arraybackslash}p{0.210\linewidth}>{\centering\arraybackslash}p{0.210\linewidth}@{}}
\caption{\textbf{Zero-shot finding annotation on PMBB abdomen contrast-enhanced (model panel 1 of 2).} Values are AUROC (95\% percentile CI) from 1,000 shared patient-cluster bootstrap resamples. Bold indicates the largest point estimate across all model panels for each finding. $n_{+}$ is the number of positive test scans.}\label{tab:supp-zs-pmbb-abd-ce-a}\\
\toprule
\textbf{Finding} & \textbf{$n_{+}$} & \textbf{ACT} & \textbf{Merlin} & \textbf{CT-CLIP} \\
\midrule
\endfirsthead
\multicolumn{5}{c}{\tablename\ \thetable\ -- continued}\\
\toprule
\textbf{Finding} & \textbf{$n_{+}$} & \textbf{ACT} & \textbf{Merlin} & \textbf{CT-CLIP} \\
\midrule
\endhead
\midrule
\multicolumn{5}{r}{Continued on next page}\\
\endfoot
\bottomrule
\endlastfoot
Hepatic Steatosis & 1424 & \mbox{0.355 (0.340, 0.371)} & \textbf{\mbox{0.629 (0.612, 0.644)}} & \mbox{0.626 (0.612, 0.640)} \\
Hepatomegaly & 313 & \mbox{0.569 (0.534, 0.603)} & \textbf{\mbox{0.639 (0.606, 0.670)}} & \mbox{0.566 (0.532, 0.600)} \\
Gallstones & 1213 & \textbf{\mbox{0.643 (0.628, 0.660)}} & \mbox{0.530 (0.513, 0.547)} & \mbox{0.560 (0.545, 0.575)} \\
Surgically Absent Gallbladder & 1767 & \mbox{0.648 (0.637, 0.661)} & \textbf{\mbox{0.862 (0.853, 0.871)}} & \mbox{0.559 (0.545, 0.571)} \\
Biliary Ductal Dilation & 514 & \mbox{0.673 (0.649, 0.694)} & \textbf{\mbox{0.764 (0.741, 0.785)}} & \mbox{0.445 (0.419, 0.468)} \\
Lymphadenopathy & 1424 & \mbox{0.569 (0.553, 0.586)} & \textbf{\mbox{0.599 (0.582, 0.616)}} & \mbox{0.531 (0.516, 0.548)} \\
Metastatic Disease & 1619 & \mbox{0.635 (0.622, 0.648)} & \mbox{0.597 (0.583, 0.612)} & \mbox{0.493 (0.480, 0.506)} \\
Renal Hypodensity & 286 & \mbox{0.559 (0.525, 0.592)} & \mbox{0.488 (0.456, 0.524)} & \mbox{0.484 (0.453, 0.515)} \\
\end{longtable}
\endgroup
\begingroup
\normalsize
\setlength{\tabcolsep}{2pt}
\renewcommand{\arraystretch}{1.0}
\setlength{\LTcapwidth}{\linewidth}
\setlength{\LTleft}{0pt}
\setlength{\LTright}{0pt plus 1fill}
\begin{longtable}{@{}>{\raggedright\arraybackslash}p{0.260\linewidth}>{\centering\arraybackslash}p{0.060\linewidth}>{\centering\arraybackslash}p{0.210\linewidth}>{\centering\arraybackslash}p{0.210\linewidth}>{\centering\arraybackslash}p{0.210\linewidth}@{}}
\caption{\textbf{Zero-shot finding annotation on PMBB abdomen contrast-enhanced (model panel 2 of 2).} Values are AUROC (95\% percentile CI) from 1,000 shared patient-cluster bootstrap resamples. Bold indicates the largest point estimate across all model panels for each finding. $n_{+}$ is the number of positive test scans.}\label{tab:supp-zs-pmbb-abd-ce-b}\\
\toprule
\textbf{Finding} & \textbf{$n_{+}$} & \textbf{MedSigLIP} & \textbf{BiomedCLIP} & \textbf{OpenAI CLIP} \\
\midrule
\endfirsthead
\multicolumn{5}{c}{\tablename\ \thetable\ -- continued}\\
\toprule
\textbf{Finding} & \textbf{$n_{+}$} & \textbf{MedSigLIP} & \textbf{BiomedCLIP} & \textbf{OpenAI CLIP} \\
\midrule
\endhead
\midrule
\multicolumn{5}{r}{Continued on next page}\\
\endfoot
\bottomrule
\endlastfoot
Hepatic Steatosis & 1424 & \mbox{0.503 (0.488, 0.517)} & \mbox{0.434 (0.419, 0.450)} & \mbox{0.577 (0.563, 0.591)} \\
Hepatomegaly & 313 & \mbox{0.498 (0.463, 0.534)} & \mbox{0.454 (0.420, 0.486)} & \mbox{0.531 (0.500, 0.561)} \\
Gallstones & 1213 & \mbox{0.560 (0.544, 0.577)} & \mbox{0.522 (0.505, 0.539)} & \mbox{0.540 (0.523, 0.555)} \\
Surgically Absent Gallbladder & 1767 & \mbox{0.559 (0.545, 0.573)} & \mbox{0.417 (0.404, 0.430)} & \mbox{0.568 (0.554, 0.581)} \\
Biliary Ductal Dilation & 514 & \mbox{0.613 (0.587, 0.638)} & \mbox{0.595 (0.569, 0.620)} & \mbox{0.465 (0.440, 0.488)} \\
Lymphadenopathy & 1424 & \mbox{0.540 (0.525, 0.556)} & \mbox{0.474 (0.459, 0.490)} & \mbox{0.538 (0.524, 0.554)} \\
Metastatic Disease & 1619 & \textbf{\mbox{0.640 (0.627, 0.653)}} & \mbox{0.533 (0.518, 0.546)} & \mbox{0.573 (0.559, 0.587)} \\
Renal Hypodensity & 286 & \textbf{\mbox{0.569 (0.531, 0.604)}} & \mbox{0.540 (0.509, 0.572)} & \mbox{0.561 (0.528, 0.597)} \\
\end{longtable}
\endgroup
\begingroup
\normalsize
\setlength{\tabcolsep}{2pt}
\renewcommand{\arraystretch}{1.0}
\setlength{\LTcapwidth}{\linewidth}
\setlength{\LTleft}{0pt}
\setlength{\LTright}{0pt plus 1fill}
\begin{longtable}{@{}>{\raggedright\arraybackslash}p{0.260\linewidth}>{\centering\arraybackslash}p{0.060\linewidth}>{\centering\arraybackslash}p{0.210\linewidth}>{\centering\arraybackslash}p{0.210\linewidth}>{\centering\arraybackslash}p{0.210\linewidth}@{}}
\caption{\textbf{Zero-shot finding annotation on RSNA-2023 abdominal trauma (model panel 1 of 2).} Values are AUROC (95\% percentile CI) from 1,000 shared patient-cluster bootstrap resamples. Bold indicates the largest point estimate across all model panels for each finding. $n_{+}$ is the number of positive test scans.}\label{tab:supp-zs-rsna2023-test-a}\\
\toprule
\textbf{Finding} & \textbf{$n_{+}$} & \textbf{ACT} & \textbf{Merlin} & \textbf{CT-CLIP} \\
\midrule
\endfirsthead
\multicolumn{5}{c}{\tablename\ \thetable\ -- continued}\\
\toprule
\textbf{Finding} & \textbf{$n_{+}$} & \textbf{ACT} & \textbf{Merlin} & \textbf{CT-CLIP} \\
\midrule
\endhead
\midrule
\multicolumn{5}{r}{Continued on next page}\\
\endfoot
\bottomrule
\endlastfoot
Low-grade kidney injury & 38 & \mbox{0.560 (0.413, 0.689)} & \mbox{0.496 (0.381, 0.607)} & \mbox{0.625 (0.540, 0.709)} \\
High-grade kidney injury & 20 & \textbf{\mbox{0.746 (0.628, 0.844)}} & \mbox{0.739 (0.604, 0.853)} & \mbox{0.542 (0.403, 0.673)} \\
Active extravasation & 61 & \mbox{0.579 (0.497, 0.656)} & \mbox{0.633 (0.552, 0.721)} & \mbox{0.552 (0.477, 0.632)} \\
Bowel injury & 24 & \textbf{\mbox{0.746 (0.599, 0.880)}} & \mbox{0.538 (0.363, 0.711)} & \mbox{0.534 (0.399, 0.677)} \\
High-grade spleen injury & 39 & \mbox{0.655 (0.566, 0.740)} & \textbf{\mbox{0.731 (0.639, 0.812)}} & \mbox{0.509 (0.387, 0.631)} \\
Low-grade spleen injury & 68 & \textbf{\mbox{0.675 (0.603, 0.745)}} & \mbox{0.631 (0.548, 0.707)} & \mbox{0.511 (0.421, 0.604)} \\
High-grade liver injury & 16 & \textbf{\mbox{0.671 (0.489, 0.834)}} & \mbox{0.574 (0.375, 0.749)} & \mbox{0.389 (0.248, 0.519)} \\
Low-grade liver injury & 84 & \textbf{\mbox{0.714 (0.645, 0.778)}} & \mbox{0.644 (0.559, 0.723)} & \mbox{0.475 (0.396, 0.561)} \\
\end{longtable}
\endgroup
\begingroup
\normalsize
\setlength{\tabcolsep}{2pt}
\renewcommand{\arraystretch}{1.0}
\setlength{\LTcapwidth}{\linewidth}
\setlength{\LTleft}{0pt}
\setlength{\LTright}{0pt plus 1fill}
\begin{longtable}{@{}>{\raggedright\arraybackslash}p{0.260\linewidth}>{\centering\arraybackslash}p{0.060\linewidth}>{\centering\arraybackslash}p{0.210\linewidth}>{\centering\arraybackslash}p{0.210\linewidth}>{\centering\arraybackslash}p{0.210\linewidth}@{}}
\caption{\textbf{Zero-shot finding annotation on RSNA-2023 abdominal trauma (model panel 2 of 2).} Values are AUROC (95\% percentile CI) from 1,000 shared patient-cluster bootstrap resamples. Bold indicates the largest point estimate across all model panels for each finding. $n_{+}$ is the number of positive test scans.}\label{tab:supp-zs-rsna2023-test-b}\\
\toprule
\textbf{Finding} & \textbf{$n_{+}$} & \textbf{MedSigLIP} & \textbf{BiomedCLIP} & \textbf{OpenAI CLIP} \\
\midrule
\endfirsthead
\multicolumn{5}{c}{\tablename\ \thetable\ -- continued}\\
\toprule
\textbf{Finding} & \textbf{$n_{+}$} & \textbf{MedSigLIP} & \textbf{BiomedCLIP} & \textbf{OpenAI CLIP} \\
\midrule
\endhead
\midrule
\multicolumn{5}{r}{Continued on next page}\\
\endfoot
\bottomrule
\endlastfoot
Low-grade kidney injury & 38 & \mbox{0.646 (0.533, 0.739)} & \mbox{0.504 (0.385, 0.616)} & \textbf{\mbox{0.651 (0.534, 0.744)}} \\
High-grade kidney injury & 20 & \mbox{0.665 (0.554, 0.768)} & \mbox{0.671 (0.490, 0.815)} & \mbox{0.436 (0.248, 0.627)} \\
Active extravasation & 61 & \textbf{\mbox{0.674 (0.605, 0.743)}} & \mbox{0.508 (0.428, 0.588)} & \mbox{0.411 (0.336, 0.490)} \\
Bowel injury & 24 & \mbox{0.664 (0.507, 0.807)} & \mbox{0.570 (0.421, 0.710)} & \mbox{0.510 (0.390, 0.636)} \\
High-grade spleen injury & 39 & \mbox{0.558 (0.443, 0.655)} & \mbox{0.487 (0.370, 0.603)} & \mbox{0.472 (0.381, 0.567)} \\
Low-grade spleen injury & 68 & \mbox{0.625 (0.527, 0.716)} & \mbox{0.537 (0.440, 0.620)} & \mbox{0.481 (0.394, 0.568)} \\
High-grade liver injury & 16 & \mbox{0.665 (0.537, 0.795)} & \mbox{0.348 (0.158, 0.536)} & \mbox{0.564 (0.392, 0.771)} \\
Low-grade liver injury & 84 & \mbox{0.550 (0.466, 0.636)} & \mbox{0.559 (0.478, 0.639)} & \mbox{0.499 (0.408, 0.583)} \\
\end{longtable}
\endgroup
 
\begin{figure}[H]
    \centering
    \includegraphics[width=\textwidth]{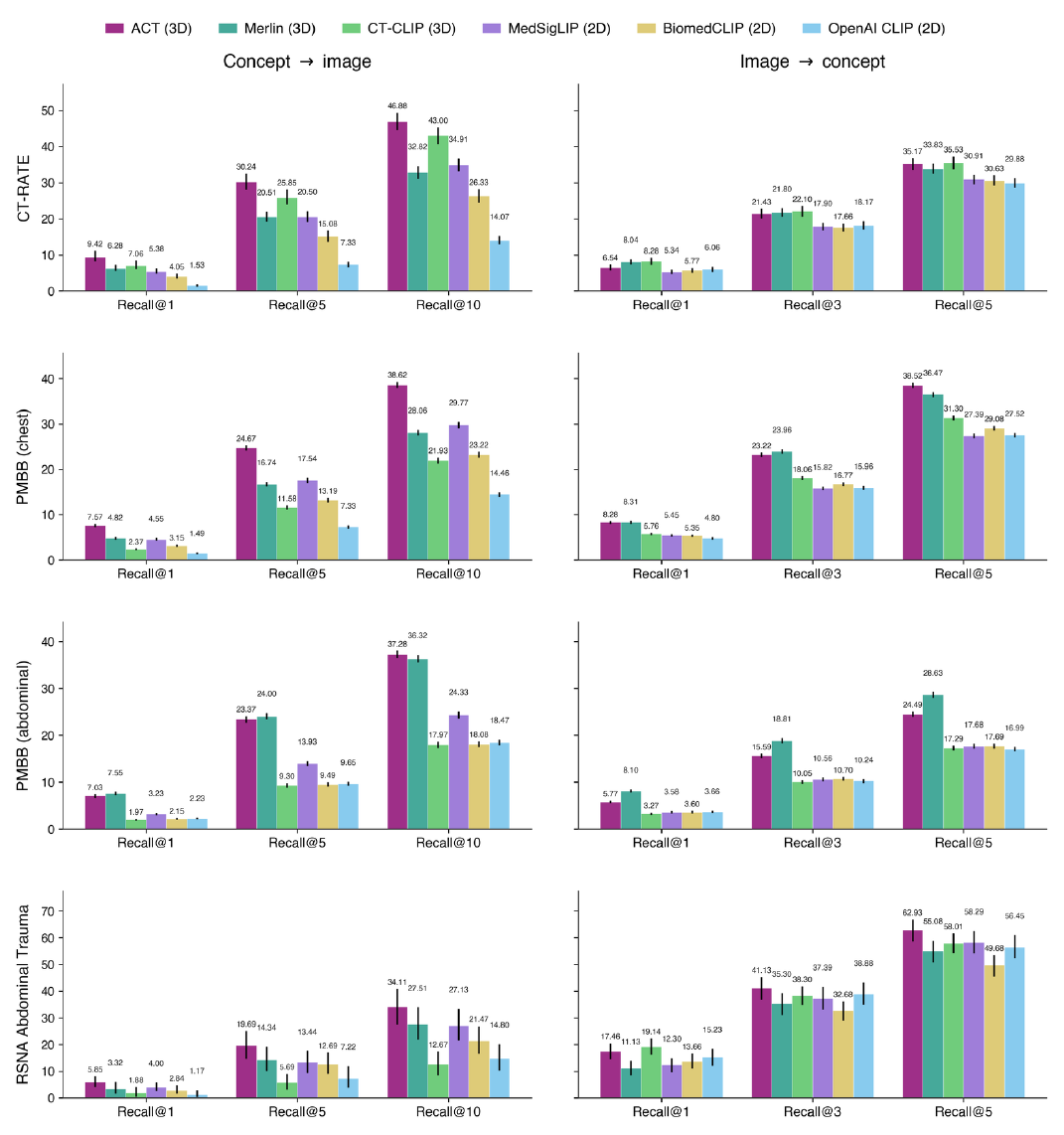}
    \caption{\textbf{Cross-modal concept retrieval with ACT's native 3D model.}
    Rows show CT-RATE, chest PMBB, abdominal PMBB and RSNA-2023, from top to bottom, for ACT's native model and five baselines.
    The left column shows concept~$\rightarrow$~image pooled Recall@1/5/10 in 64-candidate pools containing one positive and 63 negative volumes, macro-averaged over findings.
    Random floors are 1.6\%, 7.8\% and 15.6\%, respectively.
    The right column shows image~$\rightarrow$~concept Recall@1/3/5 after per-finding $z$-scoring, averaged over scans with at least one positive finding ($n$ = 2,226, 8,843, 11,646 and 265, respectively).
    Bars are full-sample point estimates.
    Error bars show percentile 95\% CIs from 1,000 patient-clustered bootstrap resamples shared across models.}
    \label{fig:retrieval}
\end{figure}

\subsection{Observation-bank composition and cross-corpus neighbours}

\begin{figure}[H]
    \centering
    \includegraphics[width=1.0\textwidth]{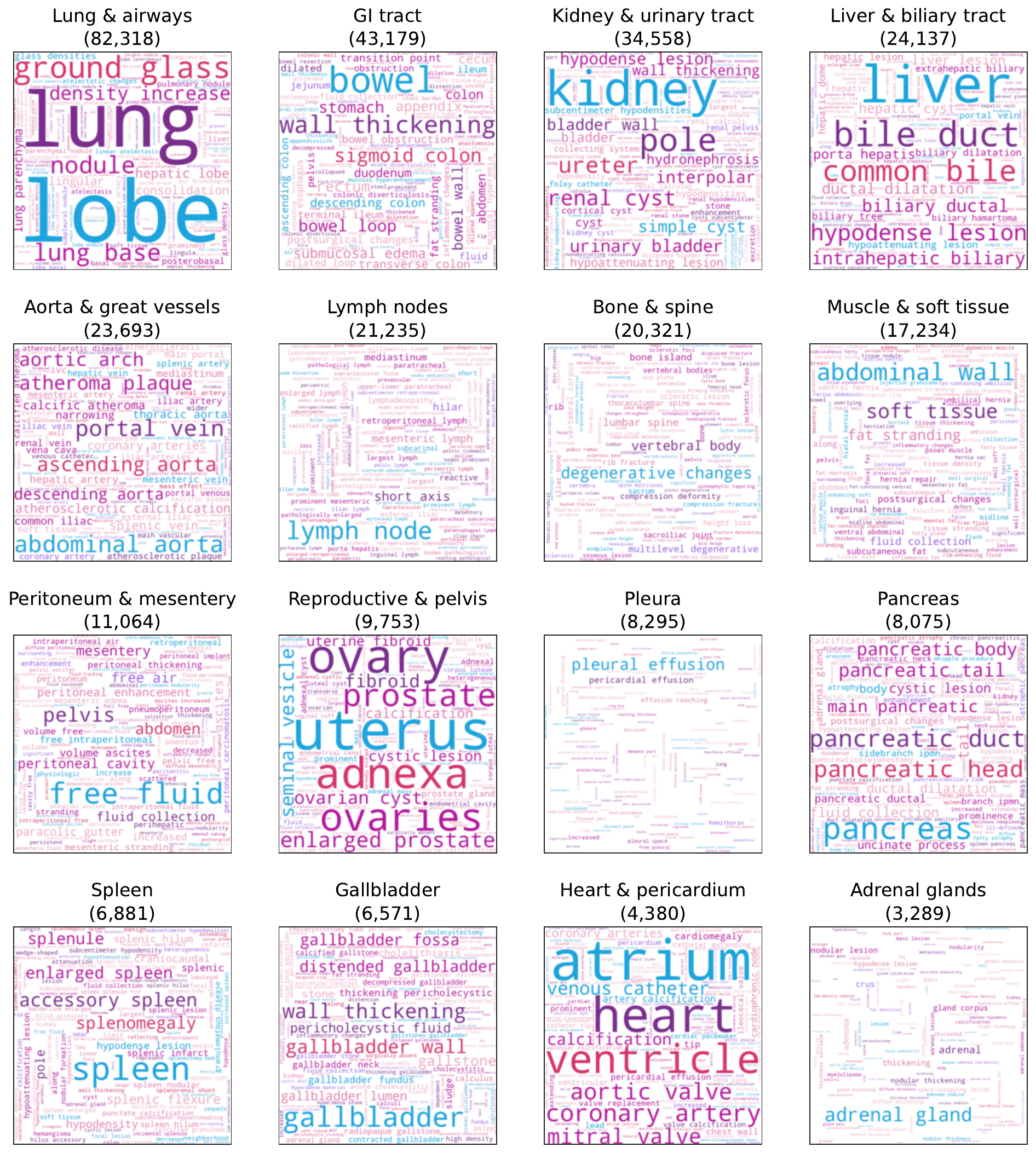}
    \caption{\textbf{Lexical content of ACT's report-derived observation bank.}
    Word clouds qualitatively summarize the 16 largest keyword-assigned categories.
    Larger words occur more frequently among the distinct observation strings.
    Category counts are shown in parentheses, colours are decorative and gastrointestinal is abbreviated as GI.}
    \label{fig:concept_wordclouds}
\end{figure}

\begingroup
\normalsize
\setlength{\tabcolsep}{2.5pt}
\renewcommand{\arraystretch}{1.0}
\setlength{\LTcapwidth}{\linewidth}
\setlength{\LTleft}{0pt}
\setlength{\LTright}{0pt plus 1fill}
\begin{longtable}{@{}L{0.160\linewidth}C{0.055\linewidth}L{0.18375\linewidth}L{0.18375\linewidth}L{0.18375\linewidth}L{0.18375\linewidth}@{}}
\caption{\textbf{Nearest affirmative and explicitly negated observations for representative PMBB queries.} The eight PMBB queries comprise four chest and four abdominal observations. Rows give the three same-finding neighbours from CT-RATE and Merlin ranked by Euclidean distance ($d$, where lower is closer) between $\ell_2$-normalized, 5,120-dimensional F2LLM observation vectors, before PCA or UMAP. $+$NN denotes an affirmative observation and $-$NN an explicit textual negation, not an independently verified negative clinical label. Exact matches, hedged statements, mixed-polarity or technical statements and phrases negating only change or degree were excluded. These selected examples are descriptive and do not estimate corpus-wide alignment accuracy.}\label{tab:pmbb-polarity-neighbors}\\
\toprule
\textbf{PMBB query} & \textbf{Rank} & \makecell[l]{\textbf{CT-RATE $+$NN}\\\textbf{affirmative}} & \makecell[l]{\textbf{CT-RATE $-$NN}\\\textbf{negated}} & \makecell[l]{\textbf{Merlin $+$NN}\\\textbf{affirmative}} & \makecell[l]{\textbf{Merlin $-$NN}\\\textbf{negated}} \\
\midrule
\endfirsthead
\multicolumn{6}{c}{\tablename\ \thetable\ -- continued}\\
\toprule
\textbf{PMBB query} & \textbf{Rank} & \makecell[l]{\textbf{CT-RATE $+$NN}\\\textbf{affirmative}} & \makecell[l]{\textbf{CT-RATE $-$NN}\\\textbf{negated}} & \makecell[l]{\textbf{Merlin $+$NN}\\\textbf{affirmative}} & \makecell[l]{\textbf{Merlin $-$NN}\\\textbf{negated}} \\
\midrule
\endhead
\midrule
\multicolumn{6}{r}{Continued on next page}\\
\endfoot
\bottomrule
\endlastfoot
\textbf{pulmonary nodules present}\newline\textit{Chest} & 1 & bilateral pulmonary nodules present\newline\mbox{$d=0.632$} & no pulmonary nodules\newline\mbox{$d=0.700$} & pulmonary nodules\newline\mbox{$d=0.539$} & no pulmonary nodules\newline\mbox{$d=0.700$} \\
 & 2 & pulmonary nodules in both lungs\newline\mbox{$d=0.705$} & no pulmonary nodule detected\newline\mbox{$d=0.803$} & numerous pulmonary nodules\newline\mbox{$d=0.740$} & no pulmonary nodules identified\newline\mbox{$d=0.734$} \\
 & 3 & pulmonary nodules are observed in both lungs\newline\mbox{$d=0.715$} & no pulmonary nodules detected in both lungs\newline\mbox{$d=0.836$} & multiple pulmonary nodules identified\newline\mbox{$d=0.741$} & no pulmonary nodule\newline\mbox{$d=0.768$} \\
\addlinespace[0.35em]\midrule
\textbf{pleural effusion}\newline\textit{Chest} & 1 & left pleural effusion\newline\mbox{$d=0.722$} & no pleural effusion\newline\mbox{$d=0.796$} & pleural effusions\newline\mbox{$d=0.553$} & no pleural effusion\newline\mbox{$d=0.796$} \\
 & 2 & new pleural effusion\newline\mbox{$d=0.735$} & no pleural effusion detected\newline\mbox{$d=0.866$} & left pleural effusion\newline\mbox{$d=0.722$} & negative for pleural effusion\newline\mbox{$d=0.841$} \\
 & 3 & pleural effusion on the left\newline\mbox{$d=0.746$} & no pleural effusion observed\newline\mbox{$d=0.880$} & pericardial pleural effusion\newline\mbox{$d=0.723$} & no pleural effusions\newline\mbox{$d=0.853$} \\
\addlinespace[0.35em]\midrule
\textbf{coronary calcification is seen}\newline\textit{Chest} & 1 & coronary artery calcifications observed\newline\mbox{$d=0.546$} & no calcific atheroma plaque in the aorta and coronary arteries\newline\mbox{$d=0.977$} & coronary calcification\newline\mbox{$d=0.527$} & no coronary artery calcification\newline\mbox{$d=0.762$} \\
\end{longtable}
\endgroup
 
\subsection{Complete CTPA-EHR phenotype performance}

\begingroup
\normalsize
\setlength{\tabcolsep}{2pt}
\renewcommand{\arraystretch}{1.0}
\setlength{\LTcapwidth}{\linewidth}
\setlength{\LTleft}{0pt}
\setlength{\LTright}{0pt plus 1fill}
\begin{longtable}{@{}>{\centering\arraybackslash}p{0.095\linewidth}>{\raggedright\arraybackslash}p{0.355\linewidth}>{\centering\arraybackslash}p{0.075\linewidth}>{\centering\arraybackslash}p{0.220\linewidth}>{\centering\arraybackslash}p{0.220\linewidth}@{}}
\caption{\textbf{Matched linear-probe performance for all 221 INSPECT phenotypes.} Values are mean AUROC (two-sided 95\% $t$-based CI) across 20 matched probe fits. Intervals quantify fit variation, not patient-sampling uncertainty. $n_{+}$ gives the positive count among 2,612 test scans, and the remaining scans are negative. Bold indicates the larger model point estimate per phenotype.}\label{tab:supp-linear-probe-221}\\
\toprule
\textbf{Phecode} & \hspace*{0.5em}\textbf{Phenotype} & \textbf{$n_{+}$} & \textbf{ACT} & \textbf{CT-CLIP} \\
\midrule
\endfirsthead
\multicolumn{5}{c}{\tablename\ \thetable\ -- continued}\\
\toprule
\textbf{Phecode} & \hspace*{0.5em}\textbf{Phenotype} & \textbf{$n_{+}$} & \textbf{ACT} & \textbf{CT-CLIP} \\
\midrule
\endhead
\midrule
\multicolumn{5}{r}{Continued on next page}\\
\endfoot
\bottomrule
\endlastfoot
165 & \hspace*{0.5em}Cancer within the respiratory system & 145 & \textbf{\mbox{0.842 (0.842, 0.843)}} & \mbox{0.722 (0.722, 0.723)} \\
165.1 & \hspace*{0.5em}Cancer of bronchus; lung & 142 & \textbf{\mbox{0.839 (0.839, 0.840)}} & \mbox{0.716 (0.715, 0.717)} \\
174 & \hspace*{0.5em}Breast cancer & 90 & \textbf{\mbox{0.627 (0.626, 0.628)}} & \mbox{0.626 (0.623, 0.628)} \\
174.1 & \hspace*{0.5em}Breast cancer [female] & 89 & \textbf{\mbox{0.625 (0.624, 0.625)}} & \mbox{0.623 (0.622, 0.625)} \\
174.11 & \hspace*{0.5em}Malignant neoplasm of female breast & 88 & \textbf{\mbox{0.627 (0.627, 0.628)}} & \mbox{0.627 (0.625, 0.629)} \\
195 & \hspace*{0.5em}Cancer, suspected or other & 179 & \textbf{\mbox{0.758 (0.758, 0.759)}} & \mbox{0.664 (0.664, 0.665)} \\
195.1 & \hspace*{0.5em}Malignant neoplasm, other & 71 & \textbf{\mbox{0.695 (0.694, 0.695)}} & \mbox{0.575 (0.574, 0.576)} \\
198 & \hspace*{0.5em}Secondary malignant neoplasm & 372 & \textbf{\mbox{0.779 (0.779, 0.779)}} & \mbox{0.678 (0.677, 0.679)} \\
\end{longtable}
\endgroup
 
\clearpage
\subsection{Complete observation-bank restriction performance}

\begingroup
\normalsize
\setlength{\tabcolsep}{2pt}
\renewcommand{\arraystretch}{1.0}
\setlength{\LTcapwidth}{\linewidth}
\setlength{\LTleft}{0pt}
\setlength{\LTright}{0pt plus 1fill}
\begin{longtable}{@{}>{\centering\arraybackslash}p{0.095\linewidth}>{\raggedright\arraybackslash}p{0.355\linewidth}>{\centering\arraybackslash}p{0.075\linewidth}>{\centering\arraybackslash}p{0.220\linewidth}>{\centering\arraybackslash}p{0.220\linewidth}@{}}
\caption{\textbf{Complete full-bank and rule-restricted ACT performance across all 86 restriction-eligible INSPECT phenotypes.} Values are held-out test AUROCs with two-sided 95\% percentile CIs from 1,000 shared patient-cluster bootstrap resamples of 2,612 scans from 2,223 patients. All scans from each sampled patient received the same multiplicity. Within each resample, AUROC was recomputed for the fixed full-bank and rule-restricted probes and summarized as one value per column. Models were not refit within resamples, so intervals quantify held-out patient-sampling rather than model-fitting variability. $n_{\mathrm{obs}}$ is the number of observations retained by the phenotype-specific rule set. Bold indicates the larger unrounded point estimate. Across all rows, mean AUROC was 0.741 for the full-bank reference and 0.751 for the rule-restricted probes, and rule-restricted values were higher for 55 phenotypes and lower for 31. No per-phenotype or aggregate hypothesis tests were conducted, and comparisons are descriptive.}\label{tab:supp-restriction-all86}\\
\toprule
\textbf{Phecode} & \hspace*{0.5em}\textbf{Phenotype} & \textbf{$n_{\mathrm{obs}}$} & \makecell{\textbf{Full-bank}\\\textbf{AUROC}\\\textbf{(95\% CI)}} & \makecell{\textbf{Rule-restricted}\\\textbf{AUROC}\\\textbf{(95\% CI)}} \\
\midrule
\endfirsthead
\multicolumn{5}{c}{\tablename\ \thetable\ -- continued}\\
\toprule
\textbf{Phecode} & \hspace*{0.5em}\textbf{Phenotype} & \textbf{$n_{\mathrm{obs}}$} & \makecell{\textbf{Full-bank}\\\textbf{AUROC}\\\textbf{(95\% CI)}} & \makecell{\textbf{Rule-restricted}\\\textbf{AUROC}\\\textbf{(95\% CI)}} \\
\midrule
\endhead
\midrule
\multicolumn{5}{r}{Continued on next page}\\
\endfoot
\bottomrule
\endlastfoot
165 & \hspace*{0.5em}Cancer within the respiratory system & \mbox{10,972} & \textbf{\mbox{0.842 (0.813, 0.870)}} & \mbox{0.823 (0.784, 0.861)} \\
165.1 & \hspace*{0.5em}Cancer of bronchus; lung & \mbox{10,972} & \textbf{\mbox{0.839 (0.808, 0.867)}} & \mbox{0.821 (0.781, 0.859)} \\
174 & \hspace*{0.5em}Breast cancer & \mbox{78} & \mbox{0.627 (0.560, 0.686)} & \textbf{\mbox{0.680 (0.612, 0.740)}} \\
174.1 & \hspace*{0.5em}Breast cancer [female] & \mbox{78} & \mbox{0.625 (0.557, 0.685)} & \textbf{\mbox{0.667 (0.602, 0.731)}} \\
174.11 & \hspace*{0.5em}Malignant neoplasm of female breast & \mbox{78} & \mbox{0.628 (0.556, 0.690)} & \textbf{\mbox{0.674 (0.606, 0.739)}} \\
195 & \hspace*{0.5em}Cancer, suspected or other & \mbox{2,479} & \mbox{0.758 (0.722, 0.791)} & \textbf{\mbox{0.775 (0.738, 0.808)}} \\
198 & \hspace*{0.5em}Secondary malignant neoplasm & \mbox{2,479} & \mbox{0.779 (0.754, 0.804)} & \textbf{\mbox{0.811 (0.787, 0.832)}} \\
198.1 & \hspace*{0.5em}Secondary malignancy of lymph nodes & \mbox{1,083} & \mbox{0.779 (0.738, 0.820)} & \textbf{\mbox{0.796 (0.757, 0.832)}} \\
\end{longtable}
\endgroup
 
\clearpage
\subsection{Probe-observation audit: discordance and nonspecific context}

\begin{figure}[H]
    \centering
    \includegraphics[width=\textwidth,height=0.94\textheight,keepaspectratio]{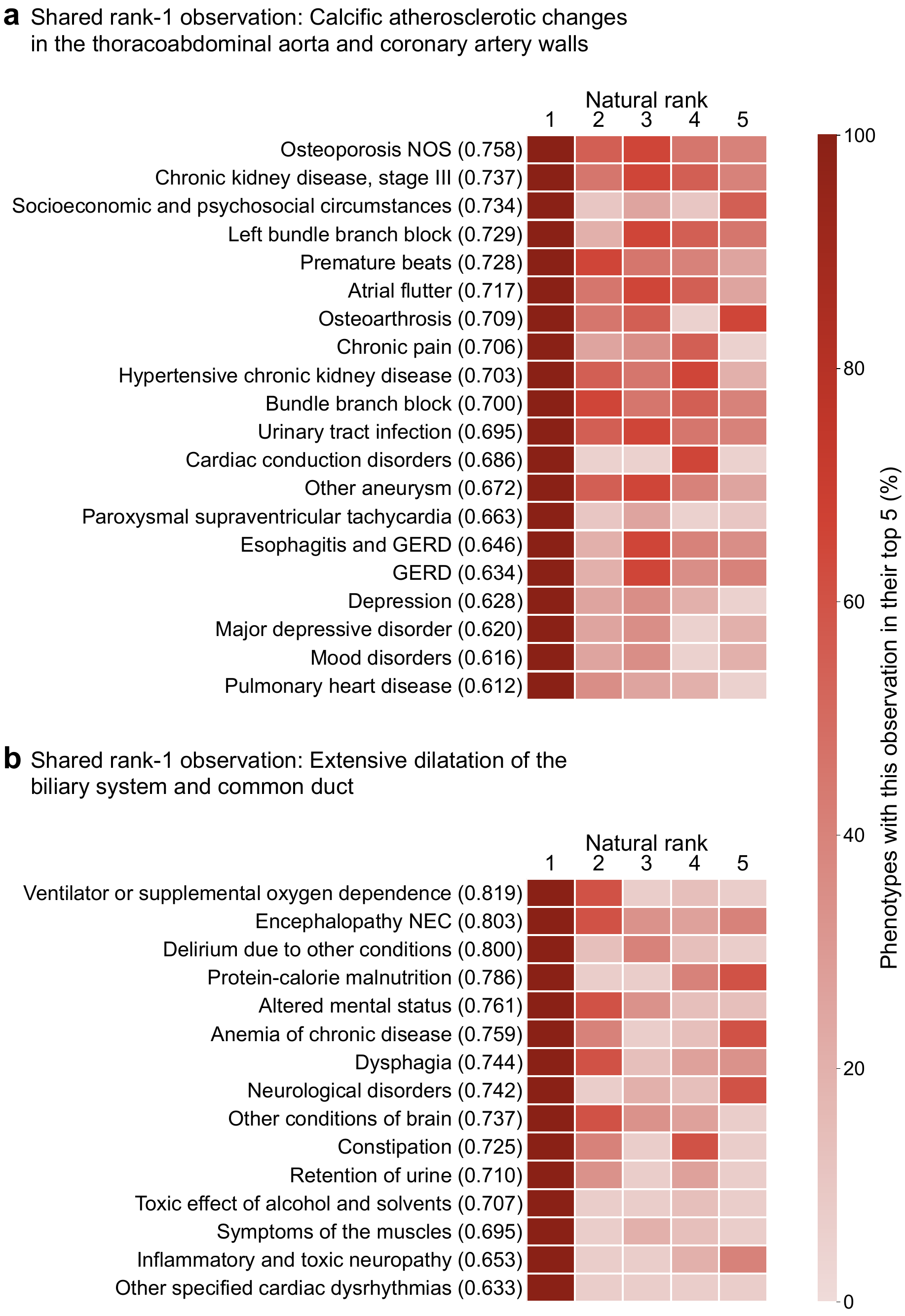}
\end{figure}

\afterpage{%
\captionof{figure}{\textbf{One observation direction leads many phenotypes that lack a common clinical target.}
    Each panel includes phenotypes sharing one top-ranked observation, named above the grid, with rows ordered by mean held-out test AUROC (in parentheses) and columns showing ranks 1 to 5 of each phenotype's own probe.
    Cell fill is the percentage of the panel's phenotypes whose own top 5 contains that cell's observation.
    \textbf{a}, The 20 phenotypes ranking ``calcific atherosclerotic changes in the thoracoabdominal aorta and coronary artery walls'' first.
    Their 100 top-5 slots hold 22 distinct observations, and members share a median of 14 of their top 25 (range 5 to 24).
    \textbf{b}, The 15 phenotypes ranking ``extensive dilatation of the biliary system and common duct'' first.
    Their 75 top-5 slots hold 34 distinct observations, and members share a median of 9 of their top 25 (range 3 to 19).
    Column 1 is 100\% by construction, since a shared rank-1 observation defines each panel.
    For columns 2 to 5, chance level under independent probe directions is 5\% in \textbf{a} and 7\% in \textbf{b}.
    Observed averages are 36\% and 23\%.
    Each panel spans 13 distinct three-digit phecode groups.
    In \textbf{a}: a socioeconomic and psychosocial circumstances code, osteoporosis NOS, urinary tract infection, and major depressive disorder.
    In \textbf{b}: constipation, retention of urine, and a toxic-exposure code.
    For each phenotype, all 376,194 bank-mean-centred, $\ell_2$-normalized F2LLM embeddings were ranked by mean dot product with the probe weight vector across 20 probe fits, without de-duplication.
    Alignment magnitudes are omitted, since unnormalized probe weights preclude comparing score magnitudes across phenotypes.
    GERD, gastro-oesophageal reflux disease. NEC, not elsewhere classified. NOS, not otherwise specified.
    \label{fig:shortcut_audit}}
\vspace{1em}
\noindent\rule{\textwidth}{0.4pt}
\vspace{1em}}
\clearpage

\begin{figure}[H]
    \centering
    \includegraphics[
        page=1,
        width=0.99\textwidth,
        height=0.62\textheight,
        keepaspectratio
    ]{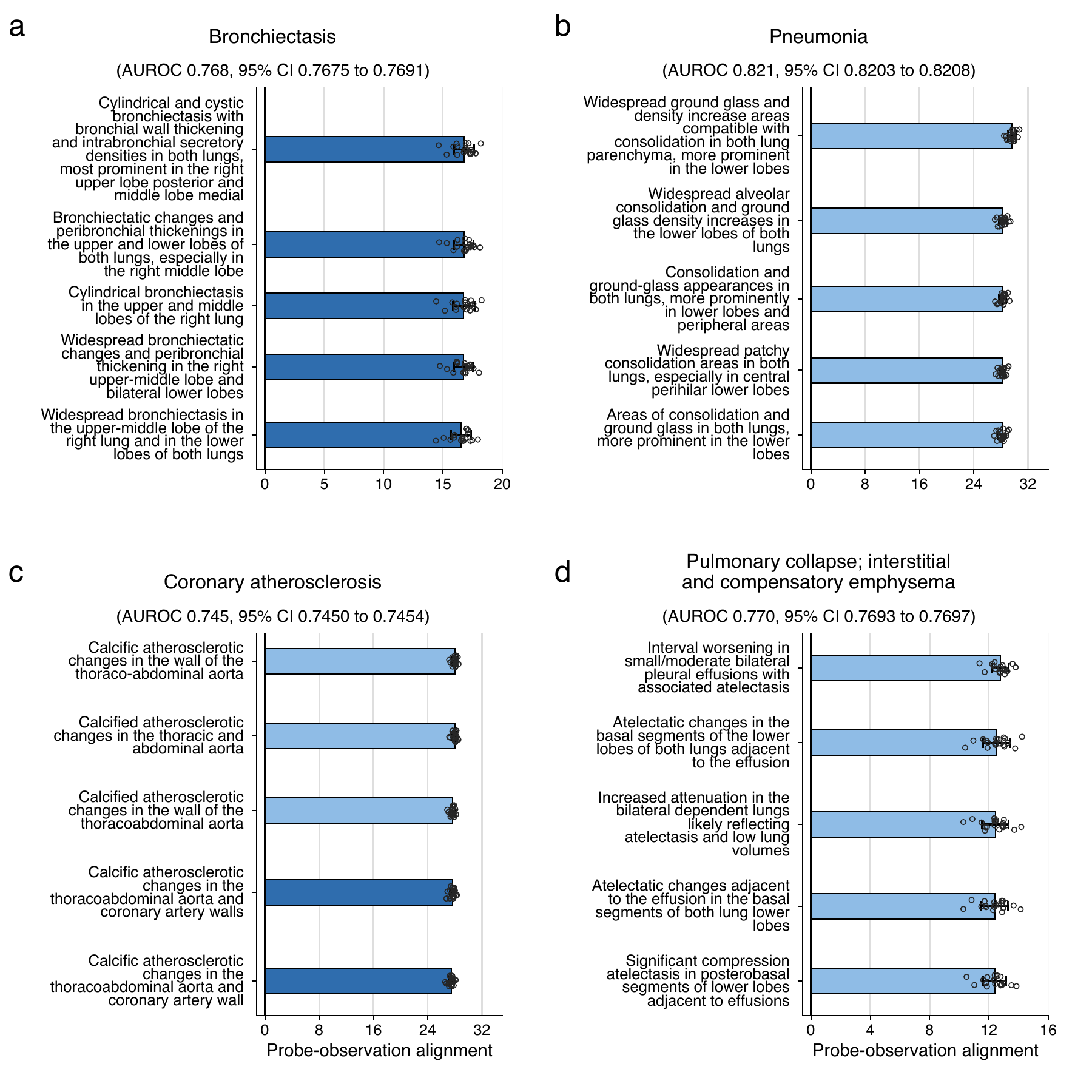}
    \caption{\textbf{Full-bank profiles distinguish direct target evidence from clinically related context.}
    \textbf{a}, Bronchiectasis.
    \textbf{b}, Pneumonia.
    \textbf{c}, Coronary atherosclerosis.
    \textbf{d}, Pulmonary collapse; interstitial and compensatory emphysema.
    Each panel shows the unmodified five highest-ranked observations for that phenotype.
    Dark blue denotes direct target evidence and light blue denotes clinically related but non-defining or qualified context.
    Colour is separate from the strict retain and reject decisions, so the uncertain but pneumonia-related rank 1 in panel \textbf{b} remains light blue.
    All bars are blue by selection, and natural full-bank ranks were neither filtered nor reranked.
    Bars show mean projections across 20 fits, data are the mean $\pm$1 s.d. and open circles show individual fits.
    Titles report mean held-out test AUROC with two-sided 95\% Student $t$ CIs across fits.
    Scales are phenotype-specific, and intervals reflect fit-to-fit rather than patient-sampling variation.}
    \label{fig:clinical_alignment_audit}
\end{figure}

\clearpage

\begin{figure}[p]
    \centering
    \includegraphics[
        width=\textwidth,
        height=0.73\textheight,
        keepaspectratio
    ]{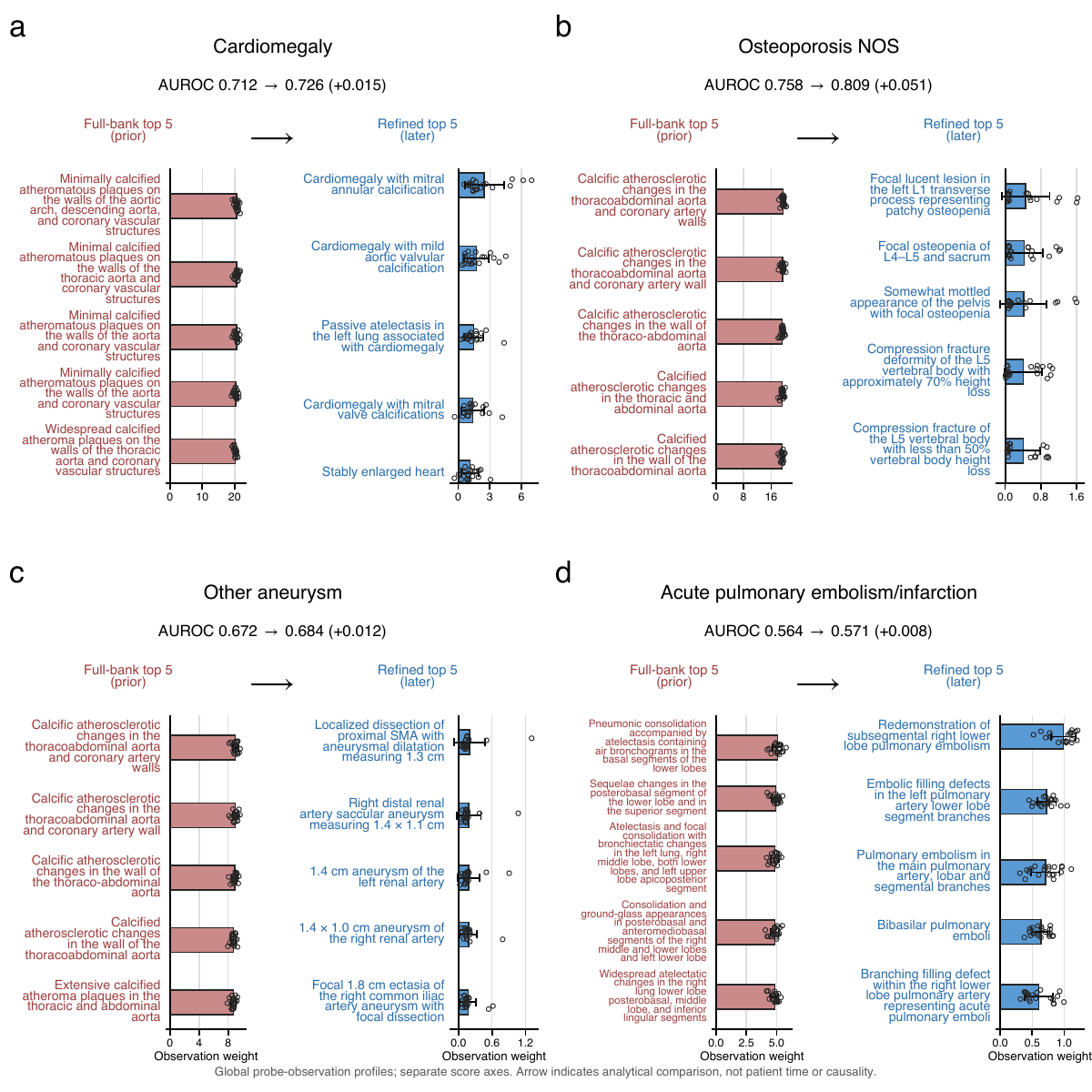}
    \caption{\textbf{Paired probe-observation audits for selected phenotypes with higher refined AUROC.}
    \textbf{a}--\textbf{d}, Paired audits for cardiomegaly (\textbf{a}), osteoporosis NOS (\textbf{b}), other aneurysm (\textbf{c}) and acute pulmonary embolism or infarction (\textbf{d}).
    Each square panel reads from the full-bank top-five audit on the left (prior) to the refined top-five audit on the right (later).
    Data are the mean $\pm$1 s.d., and open circles show individual estimates.}
    \label{fig:refinement_higher_auroc_profiles}
\end{figure}
\clearpage

\begin{figure}[p]
    \centering
    \includegraphics[
        width=\textwidth,
        height=0.73\textheight,
        keepaspectratio
    ]{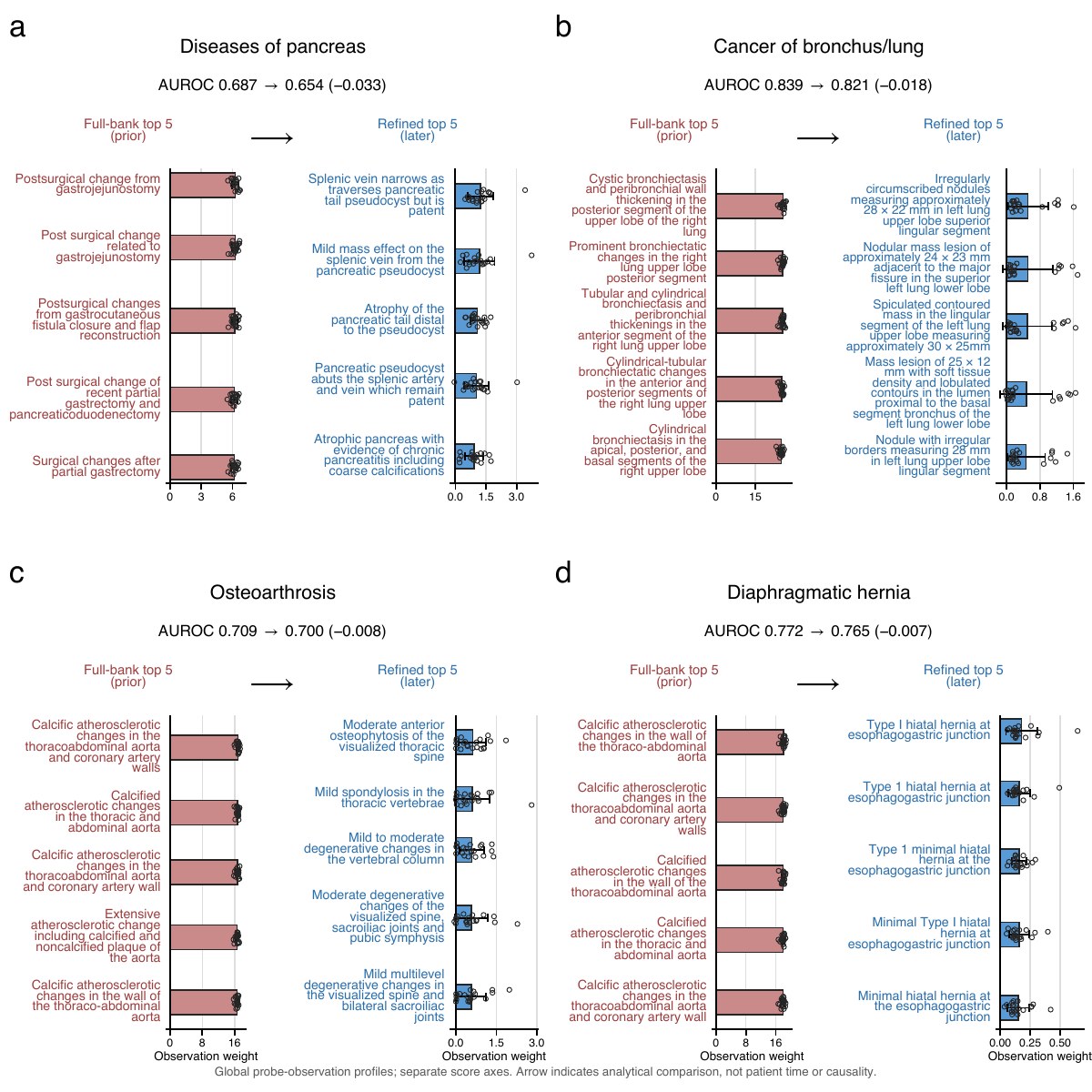}
    \caption{\textbf{Paired probe-observation audits for selected phenotypes with lower refined AUROC.}
    \textbf{a}--\textbf{d}, Paired audits for diseases of pancreas (\textbf{a}), cancer of bronchus or lung (\textbf{b}), osteoarthrosis (\textbf{c}) and diaphragmatic hernia (\textbf{d}).
    Each square panel reads from the full-bank top-five audit on the left (prior) to the refined top-five audit on the right (later).
    Data are the mean $\pm$1 s.d., and open circles show individual estimates.}
    \label{fig:refinement_lower_auroc_profiles}
\end{figure}
\clearpage

\endgroup
 
\end{document}